\documentclass{article}

    \usepackage[nonatbib, preprint]{neurips_2022}

\usepackage[utf8]{inputenc} 
\usepackage[T1]{fontenc}    
\usepackage{hyperref}       
\usepackage{url}            
\usepackage{booktabs}       
\usepackage{amsfonts}       
\usepackage{nicefrac}       
\usepackage{microtype}      
\usepackage{xcolor}         
\usepackage{amsmath}
\usepackage{amsthm}
\usepackage{array}
\usepackage{graphicx}
\usepackage{multirow}
\usepackage{subcaption}
\usepackage{caption} 
\usepackage{wrapfig}
\usepackage{hyperref}
\usepackage{colortbl}
\usepackage{bm}
\usepackage[numbers]{natbib}
\usepackage{threeparttable}

\usepackage{amssymb}
\usepackage{algorithm, algpseudocode}
\usepackage{textcomp}
\usepackage{breqn}
\usepackage{enumitem}
\usepackage{float}

\title{A Control Theoretic Framework for\\ Adaptive Gradient Optimizers in Machine Learning}

\author{%
  Kushal Chakrabarti \\
  Division of Data and Decision Sciences \\
  Tata Consultancy Services Research \\
  Mumbai 400607, India \\
  \texttt{chakrabarti.k@tcs.com} \\
  \And
  Nikhil Chopra \\
  Department of Mechanical Engineering \\
  University of Maryland \\
  College Park, Maryland 20742, U.S.A. \\
  \texttt{nchopra@umd.edu} \\
}

\def\R{\mathbb{R}}

\def\A{A^T A}

\def\A1{\left(A^1\right)^T A^1}

\providecommand{\norm}[1]{\ensuremath{\left\lVert#1\right\rVert }}
\providecommand{\mnorm}[1]{\ensuremath{\left\lvert#1\right\rvert}}

\newtheorem{theorem}{\bfseries Theorem}

\newtheorem{assumption}{\bfseries Assumption}
\newtheorem{remark}{\bfseries Remark}
\newtheorem{corollary}{\bfseries Corollary}
\newtheorem{prop}{\bfseries Proposition}

\begin{document}

\maketitle

\begin{abstract}
Adaptive gradient methods have become popular in optimizing deep neural networks; recent examples include AdaGrad and Adam. Although Adam usually converges faster, variations of Adam, for instance, the AdaBelief algorithm, have been proposed to enhance Adam's poor generalization ability compared to the classical stochastic gradient method. This paper develops a generic framework for adaptive gradient methods that solve non-convex optimization problems. We first model the adaptive gradient methods in a state-space framework, which allows us to present simpler convergence proofs of adaptive optimizers such as AdaGrad, Adam, and AdaBelief. We then utilize the transfer function paradigm from classical control theory to propose a new variant of Adam, coined AdamSSM. We add an appropriate pole-zero pair in the transfer function from squared gradients to the second moment estimate. We prove the convergence of the proposed AdamSSM algorithm. Applications on benchmark machine learning tasks of image classification using CNN architectures and language modeling using LSTM architecture demonstrate that the AdamSSM algorithm improves the gap between generalization accuracy and faster convergence than the recent adaptive gradient methods.
\end{abstract}
\section{Introduction}
\label{sec:intro}

In this paper, we consider the problem of minimizing a non-convex function. The {\em objective function} $f: \R^d \to \R_{>0}$ is smooth and, without loss of generality, positive-valued. Formally, the problem is
\begin{align}
    \min_{x \in \R^d} f(x). \label{eqn:opt_1}
\end{align}

First-order optimization algorithms, such as SGD with momentum~\cite{sutskever2013importance} and Nesterov's accelerated gradient-descent (NAG)~\cite{nesterov27method}, are popular algorithms for image classification and language modeling problems~\cite{wu2018group, luo2019learning}. These algorithms iteratively update all the model parameters using a global learning rate, and are classified as {\em accelerated gradient algorithms}.
To achieve faster convergence, several {\em adaptive gradient algorithms} have been proposed recently. These methods update each model parameter with an individual learning rate, resulting in faster training than accelerated gradient methods. AdaGrad~\cite{duchi2011adaptive} is the first prominent adaptive gradient method that significantly improved upon SGD~\cite{reddi2018convergence}. Its general version, known as G-AdaGrad~\cite{chakrabarti2021generalized}, can improve upon AdaGrad's convergence rate by tuning an additional scalar parameter. Specifically, the adaptive gradient (AdaGrad) algorithm maintains an estimate of the solution defined by~\eqref{eqn:opt_1} and updates it iteratively. The improvement over GD method is due to an adaptive updating of the learning rate, based on the information of all the past gradients. In continuous-time domain, the G-AdaGrad algorithm can be modeled using a set of ordinary differential equations (ODEs) as follows~\cite{chakrabarti2021generalized}. Let the gradient of the objective function evaluated at $x \in \R^d$ be denoted by $\nabla f(x) \in \R^d$, and its $i$-th element be denoted by $\nabla_i f(x)$ for each dimension $i \in  \{1,\ldots,d\}$. For each $i \in \{1,\ldots,d\}$ and $t \geq 0$, we consider the ODEs
\begin{align}
    \Dot{x}_{ci}(t) = \norm{\nabla_i f(x(t))}^2, \,
    \Dot{x}_i(t) = -\dfrac{\nabla_i f(x(t))}{x_{ci}(t)^c},  \label{eqn:x_g}
\end{align}
with initial conditions $\{x_{ci}(0) > 0 : i = 1,\ldots,d\}$ and $x(0) \in \R^d$. When we set the positive-valued scalar $c=0.5$,~\eqref{eqn:x_g} recover the original AdaGrad algorithm.

AdaGrad can converge slowly for dense gradients because of accumulation of all the previous gradients. A popular variant of AdaGrad is adaptive momentum estimation (Adam)~\cite{kingma2014adam}, which addresses the gradient accumulation issue in AdaGrad. In continuous-time domain, Adam can be modeled using a set of non-autonomous ODEs as follows~\cite{chakrabarti2021generalized}. We define two real-valued scalar parameters $b_1, b_2 \in (0,1)$. We further define a positive-valued function $\alpha: (0,\infty) \to \R$ by $\alpha(t) = \dfrac{1-(1-b_1)^{t+1}}{\sqrt{1-(1-b_2)^{t+1}}}, \, t\geq 0$.
For each $i \in \{1,\ldots,d\}$, we consider the following set of ODEs
\begin{align}
    \hspace{-0.2em}\Dot{\mu}_i(t) \hspace{-0.2em} = \hspace{-0.2em} -b_1 \mu_i(t) + b_1 \nabla_i f(x(t)), 
    \Dot{\nu}_i(t) \hspace{-0.2em} = \hspace{-0.2em} -b_2 \nu_i(t) + b_2 \norm{\nabla_i f(x(t))}^2, 
    \Dot{x}_i(t) \hspace{-0.2em} = \hspace{-0.2em} - \dfrac{\mu_i(t)}{\alpha(t)\sqrt{\nu_i(t)}}, \label{eqn:xm_adam}
\end{align}
with initial conditions $\mu(0) \in \R^d$, $\{\nu_i(0) > 0 : i = 1,\ldots,d\}$, and $x(0) \in \R^d$. The variables $\mu$ and $v$ are estimates of, respectively, the biased first moment and the biased second raw moment of the gradients. These variables $\mu$ and $\nu$ can be abstracted as dynamic controller states and $x$ as system state. The term $\alpha(t)$ in~\eqref{eqn:xm_adam} is responsible for the initial bias corrections in Adam~\cite{kingma2014adam}. It is known that, if the Hessian of the objective function is bounded and the parameter condition $b_1 > b_2$ holds, then Adam is guaranteed to converge to a {\em critical point} of $f$~\cite{chakrabarti2021generalized}. Note that, the algorithm parameters $b_1$ and $b_2$ in the continuous-time representation above are related to the parameters $\beta_1$ and $\beta_2$ in discrete-time implementation of Adam~\cite{kingma2014adam}, according to $\beta_1 = (1-\delta b_1)$ and $\beta_2 = (1-\delta b_2)$ where $\delta$ is the sampling-time for first-order Euler discretization of~\eqref{eqn:xm_adam}~\cite{chakrabarti2021generalized}.

Despite its widespread use, the poor generalization ability of Adam has led to its several variants. The notable ones among them include AdaBelief~\cite{NEURIPS2020_d9d4f495}, AdaBound~\cite{luo2018adaptive}, AdamW~\cite{loshchilov2018decoupled}, AMSGrad~\cite{reddi2018convergence}, Fromage~\cite{bernstein2020distance}, MSVAG~\cite{balles2018dissecting}, RAdam~\cite{liu2019variance}, and Yogi~\cite{ZaheerRSKK18}. Among these modifications of Adam, AdaBelief has been proposed most recently. The AdaBelief algorithm is similar to Adam, except that $\norm{\nabla_i f(x(t))}^2$ in~\eqref{eqn:xm_adam} is replaced by $\norm{\nabla_i f(x(t))-\mu_i(t)}^2$ so that $\frac{1}{\sqrt{\nu_i(t)}}$ represents the belief in observed gradient~\cite{NEURIPS2020_d9d4f495}. In other words, the following set of ODEs represents the AdaBelief algorithm. For each $i \in \{1,\ldots,d\}$ and $t \geq 0$,
\begin{align}
    \Dot{\mu}_i(t) & = -b_1 \mu_i(t) + b_1 \nabla_i f(x(t)), \,
    \Dot{\nu}_i(t) = -b_2 \nu_i(t) + b_2 \norm{\nabla_i f(x(t)) - \mu_i(t)}^2, \label{eqn:v_adabelief} \\
    \Dot{x}_i(t) & = - \dfrac{1}{\alpha(t)} \dfrac{\mu_i(t)}{\sqrt{\nu_i(t)}}, \label{eqn:xm_adabelief}
\end{align}
with initial conditions $\mu(0) \in \R^d$, $\{\nu_i(0) > 0 : i = 1,\ldots,d\}$, and $x(0) \in \R^d$.  Through extensive experiments, AdaBelief has been shown to reduce the generalization gap between SGD and adaptive gradient methods. However, in these experiments, few other existing methods such as Yogi, MSAVG, AdamW, and Fromage have achieved faster convergence of the training cost than AdaBelief~\cite{NEURIPS2020_d9d4f495}.

\subsection{Related work}
\label{sub:prior}

The first convergence guarantee of a generalized AdaGrad method for non-convex functions has been presented recently in~\cite{li2019convergence}, where the additional parameter $\epsilon \geq 0$ generalizes the AdaGrad method. However, the parameter $\epsilon$ in~\cite{li2019convergence} has been assumed to be strictly positive, which excludes the case of the original AdaGrad method~\cite{duchi2011adaptive} where $\epsilon = 0$. The analysis for AdaGrad in~\cite{defossez2020simple} assumes the gradients to be uniformly bounded. We do not need such an assumption. Additionally, the aforementioned analyses of AdaGrad are in discrete-time. We analyze AdaGrad in the continuous-time domain. 

Previous works that prove convergence of Adam for non-convex problems include~\cite{ZaheerRSKK18, de2018convergence, tong2019calibrating, barakat2020convergence, chen2018convergence, barakat2021convergence}. In~\cite{ZaheerRSKK18}, the proof for Adam considers the algorithm parameter $\beta_1 = 0$, which is essentially the RMSProp algorithm. We consider the general parameter settings where $\beta_1 \geq 0$. An Adam-like algorithm has been proposed and analyzed in~\cite{defossez2020simple}. The proofs in~\cite{ZaheerRSKK18, de2018convergence, tong2019calibrating, barakat2020convergence, chen2018convergence} do not consider the initial bias correction steps in the original Adam~\cite{kingma2014adam}. Our analysis of Adam considers the bias correction steps. The analyses in~\cite{defossez2020simple, ZaheerRSKK18, de2018convergence, tong2019calibrating, chen2018convergence} assume uniformly bounded gradients. We do not make such an assumption. The aforementioned analyses of Adam are in discrete-time. A continuous-time model of Adam has been proposed in~\cite{barakat2021convergence}, which includes the bias correction steps. However, compared to~\cite{barakat2021convergence}, our convergence proof for Adam is simpler. In addition,~\cite{barakat2021convergence} assumes that the parameters $\beta_1$ and $\beta_2$ in the Adam algorithm are functions of the {\em step-size} $\eta$ such that $\beta_1$ and $\beta_2$ tends to one as the {\em step-size} $\eta \to 0$. We do not make such an assumption in our analysis.  

The convergence guarantee of the AdaBelief algorithm has been provided by the authors in discrete-time~\cite{NEURIPS2020_d9d4f495}. In this paper, we present a simpler proof for AdaBelief in continuous-time.

Modeling optimization algorithms as dynamical systems is not new in the literature~\cite{su2014differential, wibisono2016variational, francca2021gradient}. ODE models of momentum-based first-order optimization algorithms have been used for analyzing their convergence properties~\cite{su2014differential, wibisono2016variational, shi2021understanding} and also for designing new algorithms~\cite{vaquero2020resource}. Recently, a unified model, refered as {\em accelerated gradient flow}, for several proximal gradient and momentum-based methods such as heavy-ball and Nesterov's method have been proposed~\cite{francca2021gradient}. Driven by this theme, we view adaptive gradient algorithms as dynamical systems in this paper, which we utilize for convergence analysis and also for proposing a novel adaptive gradient algorithm. We note that the convergence results of continuous-time optimization algorithms or {\em gradient flow} not necessarily extend to their discrete-time counterparts. In fact, different discretization schemes of the same continuous-time algorithm can result in different algorithms in discrete-time~\cite{francca2021gradient}. Moreover, most of the existing research on continuous-time modeling of optimization methods have only considered {\em accelerated gradient methods}. In this paper, we focus specifically on {\em adaptive gradient methods}.

Note that,~\cite{chakrabarti2021generalized} presented proofs from a state-space viewpoint only for G-AdaGrad and Adam. Here, we propose and analyze a state-space model that represents a class of adaptive gradient methods, from which the convergence proofs of G-AdaGrad and Adam follow as a special case. Moreover, we propose a novel adaptive gradient algorithm and conduct extensive experiments compared to~\cite{chakrabarti2021generalized}. 

\vspace{-0.6em}

\subsection{Our contributions}
\label{sub:contri}

First, we propose a class of adaptive gradient algorithms for solving non-convex optimization problems. We represent the proposed algorithm in the state-space form~\cite{khalil2015nonlinear}. This state-model of adaptive gradient algorithms is a non-autonomous system of ODEs, and includes the AdaGrad, G-AdaGrad, Adam, and AdaBelief algorithms as special cases.  We formally analyze the proposed adaptive gradient framework from a state-space perspective, which enables us in presenting simplified convergence proofs of AdaGrad, G-AdaGrad, Adam, and AdaBelief in continuous-time for non-convex optimization. Next, motivated by the issues in training neural network models as mentioned above, we propose a novel adaptive gradient algorithm for solving~\eqref{eqn:opt_1}, aimed at balancing improved generalization and faster convergence of machine learning models. We refer to the proposed algorithm as {\em Adaptive momentum estimation with Second-order dynamics of Second Moment (AdamSSM)}. Facilitated by our state-space framework of adaptive gradient methods, the proposed AdamSSM algorithm is built on top of Adam, where the dynamics of the second raw moment estimate $\nu$ of the gradients is increased in order by one.
Our key findings are summarized below.

{$\bullet$} We develop a state-space framework for a class of adaptive gradient algorithms that solves the optimization problem in~\eqref{eqn:opt_1}. Using a simple analysis of the proposed state-space model, we prove the convergence of the proposed adaptive gradient algorithm to a {\em critical point} of the possibly non-convex optimization problem~\eqref{eqn:opt_1} in the deterministic settings. Since our framework is inclusive of the AdaGrad, G-AdaGrad, Adam, and AdaBelief optimizers, an intuitive and simple convergence proof for each of these methods follows as a special case of our analysis. Please refer Section~\ref{sec:gen} for details.

{$\bullet$} We propose a new variant of Adam, which we refer as AdamSSM, for solving the general non-convex optimization problem defined in~\eqref{eqn:opt_1}. The proposed algorithm adds a specific pole-zero pair~\cite{hespanha2018linear} to the dynamics of the second raw moment estimate $\nu$ in Adam, thereby improving its convergence. Utilizing the aforementioned  framework, we guarantee convergence of the proposed AdamSSM algorithm to a {\em critical point} of~\eqref{eqn:opt_1}. Our analysis requires minimal assumptions about the optimization problem~\eqref{eqn:opt_1}. Please refer Section~\ref{sec:algo} for details.

{$\bullet$} Finally, in Section~\ref{sec:exp}, we demonstrate the applicability of our AdamSSM algorithm to benchmark machine learning problems. In this context, we conduct image classification experiments with CNN models on CIFAR10 dataset and language modeling experiments with LSTM models. These results present empirical evidence of the proposed algorithm's capability in balancing better generalization and faster convergence than the existing state-of-the-art optimizers.

\section{Generalized AdaGrad}
\label{sec:adagrad}

We first review the state-space analysis of the G-AdaGrad algorithm~\eqref{eqn:x_g}, as presented in~\cite{chakrabarti2021generalized}. This section serves as a foundation towards our general adaptive gradient algorithm proposed later in Section~\ref{sec:gen}. Detailed proofs of the theoretical guarantees in this paper are included with the supplemental material.

We make the following assumptions to present the algorithms and their convergence results. These assumptions are mild and standard in the literature of gradient-based optimization.

\begin{assumption} \label{assump_1}
Assume that the minimum of function $f$ exists and is finite, i.e., $\mnorm{\min_{x \in \R^d} f(x)} < \infty$.
\end{assumption}

\begin{assumption} \label{assump_2}
Assume that $f$ is twice differentiable over its domain $\R^d$ and the entries in the Hessian matrix $\nabla^2 f(x)$ are bounded for all $x\in \R^d$.
\end{assumption}

We define the set of {\em critical points} of the objective function $f$ as $X^* = \{x\in \R^d : \nabla f(x) = 0_d\}$.
Proposition~\ref{prop:gadagrad} below presents a key result on the convergence of G-AdaGrad~\eqref{eqn:x_g} in continuous-time to a {\em critical point} in $X^*$. 

\begin{prop}[\cite{chakrabarti2021generalized}] \label{prop:gadagrad}
Consider the pair of differential equations~\eqref{eqn:x_g} with initial conditions $x(0) \in \R^d$ and $\{x_{ci}(0) > 0 : i = 1,\ldots,d\}$. Let the parameter $c \in (0,1)$. If Assumptions~\ref{assump_1}-\ref{assump_2} hold, then $\lim_{t \to \infty} \nabla f(x(t)) = 0_d$. Moreover, for all $t \geq 0$, we have
\begin{align}
   & f(x(t)) = f(x(0)) +  \sum_{i=1}^d  \dfrac{\left(x_{ci}(0)\right)^{1-c} -  \left(x_{ci}(0) + \int_0^t \norm{\nabla_i f(x(s))}^2 ds\right)^{1-c}}{1-c}. \label{eqn:ft}
\end{align}
\end{prop}

\begin{remark}
Proposition~\ref{prop:gadagrad} relies on Barbalat's lemma~\cite{barbalat1959systemes}, a classical tool in adaptive control theory, for proving that the squared-gradient converges to zero. The first step in applying Barbalat's lemma is showing that the squared-gradient is integrable. For this, the time-derivative of the objective function $f$ is evaluated along the system trajectories. Then, both sides are integrated and substituted from the ODEs governing the controller dynamics. A simple argument then completes the first step in Barbalat's lemma. In the second step of this lemma, Assumption~\ref{assump_2} is used to show that the squared-gradient is integrable. 
The significance of this proof lies in the above steps, which guides the analysis of our general, and therefore more complex, adaptive gradient algorithm in the next section. 
\end{remark}

\section{A general adaptive gradient algorithm}
\label{sec:gen}

In this section, we propose a class of adaptive gradient algorithm, creating a state-space framework that encompasses Adam-like algorithms. We follow by presenting convergence results of some existing adaptive gradient optimizers, G-AdaGrad, Adam, and AdaBelief, from a state-space perspective.





Our framework of adaptive gradient algorithms hinges on their state-space representation in terms of ODEs. The proposed adaptive gradient algorithm is similar to Adam, described in~\eqref{eqn:xm_adam}. However, there are two notable differences. Firstly, we replace the squared-gradient term $\norm{\nabla_i f(x(t))}^2$ in the evolution of second raw moment estimate in~\eqref{eqn:xm_adam} with $\psi(\nabla_i f(x(t)), \mu_i(t))$, where we define $\psi$ as an well-behaved and non-negative valued function $\psi:\R^2 \to \R_{\geq 0}$. The sufficient conditions on the general function $\psi$ which guarantees convergence of the proposed algorithm are specified later in this section. The proposed setup includes the original Adam algorithm, i.e., $\psi(\nabla_i f(x(t)), \mu_i(t)) = \norm{\nabla_i f(x(t))}^2$, as a special case.
The other difference is that, in our algorithm, the ODE governing the second raw moment estimate of the gradients has a higher order. From simulations, we observed an issue with Adam that the ODEs in~\eqref{eqn:xm_adam} for the first and second moments can converge slowly, especially if the optimization problem~\eqref{eqn:opt_1} is ill-conditioned, leading to a slower convergence of the estimate. A similar transient property of momentum-based algorithms have been theoretically proved in~\cite{mohammadi2022transient}. Now, each of the ODEs in~\eqref{eqn:xm_adam} can be seen as a controlled dynamical system~\cite{khalil2015nonlinear}. The controlled input for $\mu_i(t)$ and $\nu_i(t)$ is, respectively, $\nabla_i f(x(t))$ and $\norm{\nabla_i f(x(t))}^2$, and the controlled output is, respectively, $\mu_i(t)$ and $\nu_i(t)$. It is known that addition of a zero improves the transient response of an linear time-invariant (LTI) system~\cite{hespanha2018linear}. Adam does not have any zero in the dynamics of $\nu_i(t)$ in~\eqref{eqn:xm_adam}. So, in our algorithm, we add a left-half plane zero to the dynamics of $\nu_i(t)$ for improving the convergence rate. However, in Adam, the dynamics of $\nu_i(t)$ is {\em strictly proper}~\cite{hespanha2018linear}. Thus, we also add an additional left half-plane pole. Hence, for each dimension $i \in \{1,\ldots,d\}$, the transfer function from $\psi(\nabla_i f(x(t)), \mu_i(t))$ to the second raw moment estimate $\nu_i(t)$ in our general setup has an additional pole-zero pair in the left-half plane, as compared to only one pole and no zero for its counterpart in Adam. Herein lies the intuition behind our algorithm.


Our general adaptive gradient algorithm is iterative wherein, at each time $t\geq 0$, it maintains four $d$-dimensional vectors: an estimate $x(t)$ of a minimum point of~\eqref{eqn:opt_1}, an estimate $\mu(t)$ of the first moment of gradients, an estimate $\nu(t)$ of the second raw moment of gradients, and an additional $\zeta(t)$ due to higher order of the transfer functions from $\psi(\nabla_i f(x(t)), \mu_i(t))$ to $\nu_i(t)$. The initial estimate $x(0)$ is chosen arbitrarily from $\R^d$. The real-valued vector variables $\mu$ and $\zeta$ are initialized at $t=0$ as the $d$-dimensional zero vector, denoted by $0_d$. The remaining variable $\nu$ is initialized according to $\{\nu_i(0) > 0 : i = 1,\ldots,d\}$. Before initiating the iterative process, the algorithm chooses nine non-negative real-valued scalar parameters $\lambda_1,\ldots,\lambda_8$ and $c$. Their specific conditions are presented later in this section. Given the parameters $\lambda_2, \lambda_6 \in (0,1)$, we define the function $\alpha_g: [0,\infty) \to \R$:
\begin{align}
    \alpha_g(t) &=
    \begin{cases}
    \dfrac{1-(1-\lambda_2)^{t+1}}{\left(1-(1-\lambda_6)^{t+1}\right)^c}, & \text{if } \lambda_7 > 0 \\
    1, &  \text{if } \lambda_7 = 0. \label{eqn:alpha_g}
    \end{cases}
\end{align}
For $t\geq 0$, the proposed adaptive gradient optimizer in continuous-time is governed by the ODEs
\begin{align}
    \Dot{\mu}_i(t) & = -\lambda_1 \mu_i(t) + \lambda_2 \nabla_i f(x(t)), \label{eqn:mu_evol} \\
    \Dot{\zeta}_i(t) & = -\lambda_3 \zeta_i(t) + \lambda_3 \nu_i(t), \label{eqn:z_evol} \\
    \Dot{\nu}_i(t) & = \lambda_4 \zeta_i(t) - \lambda_5 \nu_i(t) + \lambda_6 \psi(\nabla_i f(x(t)), \mu_i(t)), \label{eqn:nu_evol} \\
    \Dot{x}_i(t) & = - \dfrac{1}{\alpha_g(t)} \dfrac{\lambda_7 \mu_i(t) + \lambda_8 \nabla_i f(x(t))}{\nu_i(t)^c}. \label{eqn:xm_evol}
\end{align}
The $d$-dimensional vectors $\mu$, $\zeta$, and $\nu$ can be abstracted as dynamic controller states and $x$ as system state controlled by $\mu$, $\zeta$, and $\nu$. Here, $\alpha_g(t)$ represents the initial bias correction. When the moment estimate is not used in updating $x(t)$, there is no bias correction needed in the algorithm, e.g. AdaGrad~\cite{duchi2011adaptive}. Such cases are represented by $\lambda_7 = 0$ and $\alpha_g(t) = 1$ because of no bias correction.

\begin{assumption} \label{assump_3}
We assume that the function $\psi$ satisfies the following properties.
\begin{enumerate}[label={\textbf{\Alph*:}},
  ref={\theassumption.\Alph*}]
    \item $\psi$ is differentiable over $\R^2$. Moreover, $\psi(x,y)$ and its gradient $\nabla \psi (x,y)$ are bounded if both $x$ and $y$ are bounded, for any $x,y \in \R$. \label{assump_a}
    \item $\psi(x,.)$ is bounded only if $x$ is bounded, for $x \in \R$. \label{assump_b}
    \item $\lim_{t \to \infty} \psi(\nabla_i f(x(t)), \mu_i(t)) = 0$ only if $\lim_{t \to \infty} \nabla_i f(x(t)) = 0$. \label{assump_c}
\end{enumerate}
\end{assumption}

Theorem~\ref{thm:gen} below guarantees convergence of the proposed class of adaptive gradient optimizers, presented above in~\eqref{eqn:mu_evol}-\eqref{eqn:xm_evol}, to a critical point of problem~\eqref{eqn:opt_1}. 

\begin{theorem} \label{thm:gen}
Consider the set of differential equations~\eqref{eqn:mu_evol}-\eqref{eqn:xm_evol} with initial conditions $\mu(0) = \zeta(0) = 0_d$, $x(0) \in \R^d$, and $\{v_i(0) > 0 : i = 1,\ldots,d\}$. Let the parameters satisfy $0<c<1$,
\begin{align}
    & \lambda_2 > 0, \, \lambda_3 > 0, \, \lambda_4 \geq 0, \, \lambda_4 \leq \lambda_5 < 2\lambda_1/c, \, \lambda_6 > 0, \, \lambda_7 \geq 0, \, \lambda_8 \geq 0, \lambda_7+\lambda_8 > 0. \label{eqn:lambda_cond}
\end{align}
Additionally, let $0 < \lambda_6 < \lambda_2 < 1$ if $\lambda_7 > 0$.
If Assumptions~\ref{assump_1}-\ref{assump_3} hold, then $\lim_{t \to \infty} \nabla f(x(t)) = 0_d$.
\end{theorem}

Now we are ready to show that AdaGrad, G-AdaGrad, Adam, and AdaBelief are included in our proposed class of adaptive algorithms~\eqref{eqn:mu_evol}-\eqref{eqn:xm_evol}. 
G-AdaGrad in~\eqref{eqn:x_g} is a special case of~\eqref{eqn:mu_evol}-\eqref{eqn:xm_evol} with $\nu = x_c$, $\lambda_4 = \lambda_5 = 0$, $\lambda_6 = 1$, $\lambda_7 = 0$, $\lambda_8 = 1$, and $\psi(\nabla_i f(x(t)), \mu_i(t)) = \norm{\nabla_i f(x(t))}^2$.
Adam in~\eqref{eqn:xm_adam} is a special case of~\eqref{eqn:mu_evol}-\eqref{eqn:xm_evol} with $c=0.5$, $\lambda_1 = \lambda_2 = b_1$, $\lambda_4 = 0$, $\lambda_6 = b_2$, $\lambda_7 = 1$, $\lambda_8 = 0$, and $\psi(\nabla_i f(x(t)), \mu_i(t)) = \norm{\nabla_i f(x(t))}^2$.
Similarly, AdaBelief in~\eqref{eqn:v_adabelief}-\eqref{eqn:xm_adabelief} is a special case of~\eqref{eqn:mu_evol}-\eqref{eqn:xm_evol} with $c=0.5$, $\lambda_1 = \lambda_2 = b_1$, $\lambda_4 = 0$, $\lambda_6 = b_2$, $\lambda_7 = 1$, $\lambda_8 = 0$, and $\psi(\nabla_i f(x(t)), \mu_i(t)) = \norm{\nabla_i f(x(t))-\mu_i(t)}^2$. Therefore, simple convergence proofs of these algorithms for non-convex problems follow from Theorem~\ref{thm:gen}. The following results present these convergence guarantees in continuous-time. Specifically, we obtain the following corollaries of Theorem~\ref{thm:gen}.

\begin{corollary}[\bf G-AdaGrad] \label{cor:gadagrad}
Consider the set of differential equations in~\eqref{eqn:x_g} with initial conditions $x(0) \in \R^d$, and $\{x_{ci}(0) > 0 : i = 1,\ldots,d\}$. Let the parameter $c$ satisfy $0 < c < 1$. If Assumptions~\ref{assump_1}-\ref{assump_2} hold, then $\lim_{t \to \infty} \nabla f(x(t)) = 0_d$.
\end{corollary}

\begin{corollary}[\bf Adam] \label{cor:adam}
Consider the set of differential equations in~\eqref{eqn:xm_adam} with initial conditions $\mu(0) = 0_d$, $x(0) \in \R^d$, and $\{v_i(0) > 0 : i = 1,\ldots,d\}$. Let the parameters $b_1$ and $b_2$ satisfy $0 < b_2 < b_1 < 1$. If Assumptions~\ref{assump_1}-\ref{assump_2} hold, then $\lim_{t \to \infty} \nabla f(x(t)) = 0_d$.
\end{corollary}

\begin{corollary}[\bf AdaBelief] \label{cor:adabelief}
Consider the set of differential equations in~\eqref{eqn:v_adabelief}-\eqref{eqn:xm_adabelief} with initial conditions $\mu(0) = 0_d$, $x(0) \in \R^d$, and $\{v_i(0) > 0 : i = 1,\ldots,d\}$. Let the parameters $b_1$ and $b_2$ satisfy $0 < b_2 < b_1 < 1$. If Assumptions~\ref{assump_1}-\ref{assump_2} hold, then $\lim_{t \to \infty} \nabla f(x(t)) = 0_d$.
\end{corollary}

\section{Proposed algorithm: AdamSSM}
\label{sec:algo}

In this section, we present the novel AdamSSM algorithm and its convergence guarantee. The proposed AdamSSM algorithm is in the form of the generic adaptive gradient setup, presented in Section~\ref{sec:gen}, where the function $\psi$ in~\eqref{eqn:mu_evol} is defined as $\psi(\nabla_i f(x(t)), \mu_i(t)) = \norm{\nabla_i f(x(t))}^2$. Specifically, the AdamSSM algorithm in continuous-time is described by the following set of ODEs, parameterized by $b_1,b_2,$ and $b_3$. For each dimension $i \in \{1,\ldots,d\}$ and $t \geq 0$,
\begin{align}
    \Dot{\mu}_i(t) & = -b_1 \mu_i(t) + b_1 \nabla_i f(x(t)), \, \Dot{\zeta}_i(t) = -b_2 \zeta_i(t) + b_2 \nu_i(t), \label{eqn:mu_sd} \\
    \Dot{\nu}_i(t) & = b_3 \zeta_i(t) - (b_2 + b_3) \nu_i(t) + b_2 \norm{\nabla_i f(x(t))}^2, \,
    \Dot{x}_i(t) = - \dfrac{1}{\alpha(t)} \dfrac{\mu_i(t)}{\sqrt{\nu_i(t)}}. \label{eqn:xm_sd}
\end{align}

\begin{algorithm}
  \caption{AdamSSM: Adaptive momentum estimation with Second-order dynamics of Second Moment}\label{algo_1}
  \begin{algorithmic}[1]
    \State Initialize $x(0) \in \R^d$, $\mu(0) = \zeta(0) = \nu(0) =0_d$, $b_1, b_2, b_3 > 0$, $\epsilon > 0$, and the learning rate schedule $\eta(t)$.
    \For{\text{each iteration $t=0, \, 1, \, 2, \ldots $}}
      \State Compute gradient $\nabla f(x(t))$ at the current estimate.
      \State Update biased first moment estimate 
      $\mu(t+1) = (1-\delta b_1) \mu(t) + \delta b_1 \nabla f(x(t))$.
      \State Update the vector $\zeta(t)$ to 
      $\zeta(t+1) = (1-\delta b_2) \zeta(t) + \delta b_2 \nu(t)$.
      \State Update biased second raw moment estimate $\nu_i(t+1) = \delta b_3 \zeta_i(t) + (1-\delta b_2 - \delta b_3) \nu_i(t) + \delta b_2 \norm{\nabla_i f(x(t))}^2$, for each $i=1,\ldots,d$.
      \State Compute bias corrected first moment estimate
      $\Hat{\mu}(t+1) = \frac{\mu(t+1)}{1-(1-b_1)^{t+1}}$.
      \State Compute bias corrected second raw moment estimate
      $\Hat{\nu}(t+1) = \frac{\nu(t+1)}{1-(1-b_2)^{t+1}}$.
      \State Updates the current estimate $x(t)$ to
      $x_i(t+1) = x_i(t) - \eta(t) \frac{\Hat{\mu}_i(t+1)}{\sqrt{\Hat{\nu}_i(t+1)}+\epsilon}, \, i=1,\ldots,d$.
    \EndFor
  \end{algorithmic}
\end{algorithm}

The function $\psi$ and the parameters $\beta_1 = (1-\delta b_1)$ and $\beta_2 = (1-\delta b_2)$ in the proposed AdamSSM algorithm is same as Adam. However, unlike Adam, $b_3$ in AdamSSM is strictly positive. As a result, for each dimension $i \in \{1,\ldots,d\}$, the transfer function~\cite{hespanha2018linear} from $\norm{\nabla_i f(x(t))}^2$ to $\nu_i(t)$ in AdamSSM is $\frac{b_2 (s + b_2)}{s^2 + (2b_2 + b_3)s + b_2^2}$
where $s$ denotes the Laplace variable.
On the other hand, for Adam we have $b_3 = 0$, and the above transfer function $\frac{b_2}{(s + b_2)}$,
due to cancellation of a pole-zero pair at $s=-b_2$. The addition of a pole and a zero in AdamSSM is due to introduction of the positive-valued parameter $b_3$ which improves the convergence of the estimated objective function values $f(x(t))$. Note that the order of computational cost of the AdamSSM algorithm is same as Adam.

The formal convergence guarantee of the proposed AdamSSM algorithm in continuous-time is presented below in the form of Corollary~\ref{cor:AdamSSM} of Theorem~\ref{thm:gen}. 

\begin{corollary} \label{cor:AdamSSM}
Consider the set of differential equations~\eqref{eqn:mu_sd}-\eqref{eqn:xm_sd} with initial conditions $\mu(0) = \zeta(0) = 0_d$, $x(0) \in \R^d$, and $\{v_i(0) > 0 : i = 1,\ldots,d\}$. Let the parameters $b_1$, $b_2$ and $b_3$ satisfy $0 < b_2 < b_1 < 1, \, b_3 > 0, \, b_2 + b_3 < 4b_1$.
If Assumptions~\ref{assump_1}-\ref{assump_2} hold, then $\lim_{t \to \infty} \nabla f(x(t)) = 0_d$.
\end{corollary}

The AdamSSM algorithm in discrete-time is summarized above in Algorithm~\ref{algo_1}. Specifically, we use first-order {\em explicit} Euler discretization with a fixed sampling rate to discretize the set of ODEs~\eqref{eqn:mu_sd}-\eqref{eqn:xm_sd}. To be consistent with the format of the existing algorithms, we replace the condition $\nu_i(0) > 0 \, \forall i$ with an additional parameter $\epsilon > 0$, as it is in the existing algorithms. The purpose of either of these conditions is the same: to avoid division by zero in~\eqref{eqn:xm_sd}. Additionally, we denote the learning rate parameter for updating the estimate $x(t)$ for each iteration $t = 0,1,\ldots$ in discrete-time by $\eta(t)$.

\section{Experimental results}
\label{sec:exp}

In this section, we present experimental results on benchmark machine learning problems, comparing the convergence rate and test-set accuracy of the proposed AdamSSM algorithm with several other adaptive gradient methods: AdaBelief~\cite{NEURIPS2020_d9d4f495}, AdaBound~\cite{luo2018adaptive}, Adam~\cite{kingma2014adam}, AdamW~\cite{loshchilov2018decoupled}, Fromage~\cite{bernstein2020distance}, MSVAG~\cite{balles2018dissecting}, RAdam~\cite{liu2019variance}, SGD~\cite{bottou2018optimization}, and Yogi~\cite{ZaheerRSKK18}. In the experiments, we consider two machine learning tasks: image classification on CIFAR-10 dataset~\cite{krizhevsky2009learning} and language modeling on Penn TreeBank (PTB) dataset~\cite{marcus1993building}. For image classification task, we use two CNN architectures: ResNet34~\cite{he2016deep} and VGG11~\cite{simonyan2014very}. For language modeling task, we use the long short-term memory (LSTM)~\cite{ma2015long} architecture with respectively 1-layer, 2-layers, and 3-layers. Additional details on these experiments and related figures are in the supplemental material. 

To conduct the experiments, we adapt the experimental setup used in the recent AdaBelief paper~\cite{NEURIPS2020_d9d4f495} and the AdaBound paper~\cite{luo2018adaptive}. Following~\cite{NEURIPS2020_d9d4f495}, the $l_2$-regularization hyperparameter is set to $5 \times 10^{-4}$ for image classification and $1.2 \times 10^{-6}$ for language modeling, for each algorithm. The hyperparameters are tuned such that the individual algorithms achieves a better generalization on the test set. Following~\cite{NEURIPS2020_d9d4f495}, these hyperparameters are selected as described below. 

{\bf AdaBelief}: The standard parameter values $\beta_1 = 0.9$, $\beta_2 = 0.999$ are used. $\epsilon$ is set to $10^{-8}$ for image classification, $10^{-16}$ for 1-layer LSTM and $10^{-12}$ for 2-layer and 3-layer LSTM. The learning rate $\eta$ is set to $10^{-2}$ for 2-layer and 3-layer LSTM, and $10^{-3}$ for all other models. These parameter values are set according to the implementation of AdaBelief in GitHub~\cite{AdaBelief}.
{\bf AdaBound, Adam, MSVAG, RAdam, Yogi}: The parameter $\beta_1$ is selected from $\{0.5,0.6,0.7,0.8,0.9\}$. The learning rate $\eta$ is selected from $\{10^p: \, p = 1,0,-1,-2,-3\}$. Standard values are used for the other parameters. 
{\bf AdamW}: The weight decay parameter is chosen from $\{10^{-2}, 10^{-3}, 5 \times 10^{-4}, 10^{-4}\}$. The other parameters are selected in the same way as Adam.
{\bf Fromage, SGD}: The learning rate is selected as described above for Adam. The momentum is chosen as the default value of $0.9$.
{\bf AdamSSM}: The parameters $\beta_1 = (1-\delta b_1)$ and $\beta_2 = (1-\delta b_2)$ in the proposed AdamSSM algorithm is similar to Adam. The standard choices for $\beta_1$ and $\beta_2$ in Adam are respectively $0.9$ and $0.999$~\cite{kingma2014adam}. With a sampling time $\delta = 0.15$, therefore we set $b_1 = 0.67$ and $b_2 = 0.0067$. The parameter $b_3$ is chosen from $\{c\times 10^{-3}/\delta: c = 1,2,3,4,5\}$. The parameter $\epsilon$ and the learning rate $\eta$ are same as AdaBelief.

For the image classification tasks, the model is trained for $200$ epochs; the learning rate is multiplied by $0.1$ at epoch $150$; and a mini-batch size of $128$ is used~\cite{NEURIPS2020_d9d4f495, luo2018adaptive}. We compare the training-set and test-set accuracy of different training algorithms in Table~\ref{tab:resnet} and Table~\ref{tab:vgg}. We observe that the proposed AdamSSM algorithm has the best test-set accuracy among all the algorithms, on both the architectures ResNet34 and VGG11. Some other algorithms achieve a better training-set accuracy than the proposed method. However, the test-set accuracy of those algorithms is less than AdamSSM.

For the language modeling tasks, the model is trained for $200$ epochs; the learning rate is multiplied by $0.1$ at epoch $100$ and $145$; and a mini-batch size of $20$ is used~\cite{NEURIPS2020_d9d4f495}. We compare the training-set and test-set perplexity of different training algorithms in Table~\ref{tab:lstm1}-\ref{tab:lstm3}. Note that a lower perplexity means better accuracy. For 1-layer LSTM, only the Adam method generates lower test set perplexity than the proposed method. For 2-layer LSTM, only the AdaBelief method generates lower test set perplexity than the proposed method. For the more complex 3-layer LSTM, the proposed method achieves both the least test set and the least training set perplexity.

\begin{table*}[htb!]
\caption{Comparisons between best training accuracy, best test accuracy, and number of training epochs required to achieve these accuracies for different algorithms on image classification task with ResNet34.}
\begin{center}
\begin{tabular}{|c||c|c||c|c|}
\hline
Training algorithm & Test accuracy & Epoch & Train accuracy & Epoch \\
\hline
\hline
AdaBelief & $95.44$ & $194$ & $99.988$ & $193$ \\
\hline
AdaBound & $94.85$ & $190$ & $99.998$ & $191$ \\
\hline
Adam & $93.02$ & $189$ & $99.308$ & $190$ \\
\hline
AdamW & $94.59$ & $164$ & ${\bf 100.0}$ & $169$ \\
\hline
Fromage & $94.51$ & $165$ & $99.992$ & $165$ \\
\hline
MSVAG & $94.44$ & $199$ & $99.996$ & $185$ \\
\hline
RAdam & $94.33$ & $182$ & ${\bf 100.0}$ & $179$ \\
\hline
SGD & $94.64$ & $155$ & $99.272$ & $169$ \\
\hline
Yogi & $94.71$ & $182$ & $99.972$ & $192$ \\
\hline
AdamSSM (Proposed) & ${\bf 95.61}$ & $174$ & $99.99$ & $188$ \\
\hline
\end{tabular}
\end{center}
\label{tab:resnet}
\end{table*}

\begin{table*}[htb!]
\caption{Comparisons between best training accuracy, best test accuracy, and number of training epochs required to achieve these accuracies for different algorithms on image classification task with VGG11.}
\begin{center}
\begin{tabular}{|c||c|c||c|c|}
\hline
Training algorithm & Test accuracy & Epoch & Train accuracy & Epoch \\
\hline
\hline
AdaBelief & $91.41$ & $193$ & $99.784$ & $197$ \\
\hline
AdaBound & $90.62$ & $176$ & $99.914$ & $193$\\
\hline
Adam & $88.40$ & $197$ & $94.028$ & $199$ \\
\hline
AdamW & $89.39$ & $166$ & $99.312$ & $198$ \\
\hline
Fromage & $89.77$ & $162$ & $99.730$ & $170$ \\
\hline
MSVAG & $90.24$ & $187$ & ${\bf 99.948}$ & $192$ \\
\hline
RAdam & $89.30$ & $195$ & $98.984$ & $196$ \\
\hline
SGD & $90.11$ & $188$ & $96.436$ & $195$\\
\hline
Yogi & $90.67$ & $192$ & $99.868$ & $196$ \\
\hline
AdamSSM (Proposed) & ${\bf 91.49}$ & $185$ & $99.792$ & $187$ \\
\hline
\end{tabular}
\end{center}
\label{tab:vgg}
\end{table*}

\begin{table*}[htb!]
\caption{Comparisons between best training set perplexity, best test set perplexity, and number of training epochs required to achieve these perplexities for different algorithms on language modeling task with 1-layer LSTM.}
\begin{center}
\begin{tabular}{|c||c|c||c|c|}
\hline
Training algorithm & Test accuracy & Epoch & Train accuracy & Epoch \\
\hline
\hline
AdaBelief & $84.63$ & $199$ & $58.25$ & $192$ \\
\hline
AdaBound & $84.78$ & $199$ & $62.36$ & $155$ \\
\hline
Adam & ${\bf 84.28}$ & $196$ & $58.26$ & $155$ \\
\hline
AdamW & $87.80$ & $194$ & ${\bf 55.33}$ & $155$ \\
\hline
MSVAG & $84.68$ & $199$ & $63.59$ & $167$ \\
\hline
RAdam & $88.57$ & $196$ & $55.81$ & $155$ \\
\hline
SGD & $85.07$ & $199$ & $63.64$ & $155$ \\
\hline
Yogi & $86.59$ & $199$ & $69.22$ & $155$ \\
\hline
AdamSSM (Proposed) & $84.61$ & $199$ & $58.93$ & $192$ \\
\hline
\end{tabular}
\end{center}
\label{tab:lstm1}
\end{table*}

\begin{table*}[htb!]
\caption{Comparisons between best training set perplexity, best test set perplexity, and number of training epochs required to achieve these perplexities for different algorithms on language modeling task with 2-layer LSTM.}
\begin{center}
\begin{tabular}{|c||c|c||c|c|}
\hline
Training algorithm & Test accuracy & Epoch & Train accuracy & Epoch \\
\hline
\hline
AdaBelief & ${\bf 66.29}$ & $199$ & $45.48$ & $184$ \\
\hline
AdaBound & $67.53$ & $199$ & ${\bf 43.65}$ & $165$ \\
\hline
Adam & $67.27$ & $199$ & $46.86$ & $184$ \\
\hline
AdamW & $94.86$ & $186$ & $67.51$ & $184$ \\
\hline
MSVAG & $68.84$ & $199$ & $45.90$ & $184$ \\
\hline
RAdam & $90.00$ & $199$ & $61.48$ & $184$ \\
\hline
SGD & $67.42$ & $197$ & $44.79$ & $165$ \\
\hline
Yogi & $71.33$ & $199$ & $54.53$ & $143$ \\
\hline
AdamSSM (Proposed) & $66.75$ & $198$ & $44.92$ & $190$ \\
\hline
\end{tabular}
\end{center}
\label{tab:lstm2}
\end{table*}

\begin{table*}[htb!]
\caption{Comparisons between best training set perplexity, best test set perplexity, and number of training epochs required to achieve these perplexities for different algorithms on language modeling task with 3-layer LSTM.}
\begin{center}
\begin{tabular}{|c||c|c||c|c|}
\hline
Training algorithm & Test accuracy & Epoch & Train accuracy & Epoch \\
\hline
\hline
AdaBelief & $61.24$ & $194$ & $37.06$ & $197$ \\
\hline
AdaBound & $63.58$ & $195$ & $37.85$ & $193$ \\
\hline
Adam & $64.28$ & $199$ & $43.11$ & $197$ \\
\hline
AdamW & $104.49$ & $159$ & $104.94$ & $155$ \\
\hline
MSVAG & $65.04$ & $192$ & $39.64$ & $185$ \\
\hline
RAdam & $93.11$ & $199$ & $90.75$ & $185$ \\
\hline
SGD & $63.77$ & $146$ & $38.11$ & $146$ \\
\hline
Yogi & $67.51$ & $196$ & $51.46$ & $164$ \\
\hline
AdamSSM (Proposed) & ${\bf 61.18}$ & $188$ & ${\bf 36.82}$ & $197$ \\
\hline
\end{tabular}
\end{center}
\label{tab:lstm3}
\end{table*}
\section{Discussion and conclusions}
\label{sec:summary}

 This paper proposed an adaptive gradient-descent optimizer to accelerate gradient-based optimization of non-convex optimization problems. Adaptive gradient optimizers are an integral component of modern-day machine learning pipelines. We presented a state-space framework for such optimizers, facilitating a dynamical system perspective of some prominent adaptive gradient optimizers and their potential variations. Specifically, the estimate of the true minima is the system state, and updating the learning rate of the state can be described as a control input. When the learning rates are updated adaptively, the control input is dynamically evolving. Hence, the dynamic controller states are governed by feedback inputs that are a function of the current estimate (the current system state). The existing optimizers specify the feedback input to update the dynamic controller states. We proposed a general class of feedback inputs to the controller states and specified sufficient conditions for the feedback to guarantee convergence of our general algorithm for adaptive optimization. Another salient feature of our generic algorithm is the addition of a suitable pole and zero in the transfer function to the controller state from its corresponding input. By rigorously analyzing this state-space model of the proposed algorithm, we developed a framework from which simplified proofs of existing algorithms, such as G-AdaGrad, Adam, and AdaBelief, follow. This framework can yield constructive insights to design new optimizers, as we have shown by developing the AdamSSM algorithm. AdamSSM is a variant of Adam, where adding a pole-zero pair in the second-moment estimate of gradients improves the convergence through a re-balance of transient and steady state. Through extensive experiments on complex machine learning problems, we demonstrated that the AdamSSM algorithm reduces the gap between better accuracy on unseen data and faster convergence than the state-of-the-art algorithms. Further intuition behind the AdamSSM algorithm's ability to improve generalization error is presented in the supplemental material with numerical evidence (ref. Figure~\ref{fig:nu}).

A new optimizer can also be developed by modifying the AdaBelief algorithm similarly. Formally speaking, this new variant can be described by our generic state-space framework~\eqref{eqn:mu_evol}-\eqref{eqn:xm_evol} as follows:
\begin{align*}
    \Dot{\mu}_i(t) & = -b_1 \mu_i(t) + b_1 \nabla_i f(x(t)),
    \Dot{\zeta}_i(t) = -b_2 \zeta_i(t) + b_2 \nu_i(t), \\
    \Dot{\nu}_i(t) & = b_3 \zeta_i(t) - (b_2 + b_3) \nu_i(t) + b_2 \norm{\nabla_i f(x(t)) - \mu_i(t)}^2,
    \Dot{x}_i(t) = - \dfrac{1}{\alpha(t)} \dfrac{\mu_i(t)}{\sqrt{\nu_i(t)}}. 
\end{align*}
Convergence of the above algorithm directly follows from Theorem~\ref{thm:gen}. However, further studies are needed to validate the efficacy of this optimizer compared to the existing optimizers.
In this work, we modified the transfer function only for the second moment estimate of the gradients. The idea can be extended to modifying the transfer function for the first moment estimate as well. Additionally, other input functions $\psi$ that satisfy Assumption~\ref{assump_3} can be explored for developing potentially better optimizers. In this work, we have analyzed asymptotic convergence. Another direction of work is characterizing the (finite-time) rate of convergence.

\newpage
\bibliography{refs}

\begin{thebibliography}{40}
\providecommand{\natexlab}[1]{#1}
\providecommand{\url}[1]{\texttt{#1}}
\expandafter\ifx\csname urlstyle\endcsname\relax
  \providecommand{\doi}[1]{doi: #1}\else
  \providecommand{\doi}{doi: \begingroup \urlstyle{rm}\Url}\fi

\bibitem[Sutskever et~al.(2013)Sutskever, Martens, Dahl, and
  Hinton]{sutskever2013importance}
Ilya Sutskever, James Martens, George Dahl, and Geoffrey Hinton.
\newblock On the importance of initialization and momentum in deep learning.
\newblock In \emph{International conference on machine learning}, pages
  1139--1147. PMLR, 2013.

\bibitem[Nesterov(1983)]{nesterov27method}
Y~Nesterov.
\newblock A method of solving a convex programming problem with convergence
  rate {$O$}($1/k^2$).
\newblock \emph{Sov. Math. Doklady}, 27\penalty0 (2):\penalty0 372--376, 1983.

\bibitem[Wu and He(2018)]{wu2018group}
Yuxin Wu and Kaiming He.
\newblock Group normalization.
\newblock In \emph{Proceedings of the European conference on computer vision
  (ECCV)}, pages 3--19, 2018.

\bibitem[Luo et~al.(2019{\natexlab{a}})Luo, Huang, Zeng, Nie, and
  Sun]{luo2019learning}
Liangchen Luo, Wenhao Huang, Qi~Zeng, Zaiqing Nie, and Xu~Sun.
\newblock Learning personalized end-to-end goal-oriented dialog.
\newblock In \emph{Proceedings of the AAAI Conference on Artificial
  Intelligence}, volume~33, pages 6794--6801, 2019{\natexlab{a}}.

\bibitem[Duchi et~al.(2011)Duchi, Hazan, and Singer]{duchi2011adaptive}
John Duchi, Elad Hazan, and Yoram Singer.
\newblock Adaptive subgradient methods for online learning and stochastic
  optimization.
\newblock \emph{Journal of machine learning research}, 12\penalty0 (7), 2011.

\bibitem[Reddi et~al.(2018)Reddi, Kale, and Kumar]{reddi2018convergence}
Sashank~J Reddi, Satyen Kale, and Sanjiv Kumar.
\newblock On the convergence of adam and beyond.
\newblock In \emph{International Conference on Learning Representations}, 2018.

\bibitem[Chakrabarti and Chopra(2021)]{chakrabarti2021generalized}
Kushal Chakrabarti and Nikhil Chopra.
\newblock Generalized {A}da{G}rad {(G-AdaGrad)} and {A}dam: A state-space
  perspective.
\newblock In \emph{2021 60th IEEE Conference on Decision and Control (CDC)}.
  IEEE, 2021.
\newblock Accepted.

\bibitem[Kingma and Ba(2014)]{kingma2014adam}
Diederik~P Kingma and Jimmy Ba.
\newblock Adam: A method for stochastic optimization.
\newblock \emph{arXiv preprint arXiv:1412.6980}, 2014.

\bibitem[Zhuang et~al.(2020)Zhuang, Tang, Ding, Tatikonda, Dvornek,
  Papademetris, and Duncan]{NEURIPS2020_d9d4f495}
Juntang Zhuang, Tommy Tang, Yifan Ding, Sekhar~C Tatikonda, Nicha Dvornek,
  Xenophon Papademetris, and James Duncan.
\newblock Adabelief optimizer: Adapting stepsizes by the belief in observed
  gradients.
\newblock In \emph{Advances in Neural Information Processing Systems},
  volume~33, pages 18795--18806, 2020.

\bibitem[Luo et~al.(2019{\natexlab{b}})Luo, Xiong, Liu, and
  Sun]{luo2018adaptive}
Liangchen Luo, Yuanhao Xiong, Yan Liu, and Xu~Sun.
\newblock Adaptive gradient methods with dynamic bound of learning rate.
\newblock In \emph{International Conference on Learning Representations},
  2019{\natexlab{b}}.

\bibitem[Loshchilov and Hutter(2019)]{loshchilov2018decoupled}
Ilya Loshchilov and Frank Hutter.
\newblock Decoupled weight decay regularization.
\newblock In \emph{International Conference on Learning Representations}, 2019.

\bibitem[Bernstein et~al.(2020)Bernstein, Vahdat, Yue, and
  Liu]{bernstein2020distance}
Jeremy Bernstein, Arash Vahdat, Yisong Yue, and Ming-Yu Liu.
\newblock On the distance between two neural networks and the stability of
  learning.
\newblock \emph{arXiv preprint arXiv:2002.03432}, 2020.

\bibitem[Balles and Hennig(2018)]{balles2018dissecting}
Lukas Balles and Philipp Hennig.
\newblock Dissecting adam: The sign, magnitude and variance of stochastic
  gradients.
\newblock In \emph{International Conference on Machine Learning}, pages
  404--413. PMLR, 2018.

\bibitem[Liu et~al.(2020)Liu, Jiang, He, Chen, Liu, Gao, and
  Han]{liu2019variance}
Liyuan Liu, Haoming Jiang, Pengcheng He, Weizhu Chen, Xiaodong Liu, Jianfeng
  Gao, and Jiawei Han.
\newblock On the variance of the adaptive learning rate and beyond.
\newblock In \emph{International Conference on Learning Representations}, 2020.

\bibitem[Zaheer et~al.(2018)Zaheer, Reddi, Sachan, Kale, and
  Kumar]{ZaheerRSKK18}
Manzil Zaheer, Sashank~J. Reddi, Devendra~Singh Sachan, Satyen Kale, and Sanjiv
  Kumar.
\newblock Adaptive methods for nonconvex optimization.
\newblock In \emph{NeurIPS}, pages 9815--9825, 2018.

\bibitem[Li and Orabona(2019)]{li2019convergence}
Xiaoyu Li and Francesco Orabona.
\newblock On the convergence of stochastic gradient descent with adaptive
  stepsizes.
\newblock In \emph{The 22nd International Conference on Artificial Intelligence
  and Statistics}, pages 983--992. PMLR, 2019.

\bibitem[D{\'e}fossez et~al.(2020)D{\'e}fossez, Bottou, Bach, and
  Usunier]{defossez2020simple}
Alexandre D{\'e}fossez, L{\'e}on Bottou, Francis Bach, and Nicolas Usunier.
\newblock A simple convergence proof of {A}dam and {A}dagrad.
\newblock \emph{arXiv preprint arXiv:2003.02395}, 2020.

\bibitem[De et~al.(2018)De, Mukherjee, and Ullah]{de2018convergence}
Soham De, Anirbit Mukherjee, and Enayat Ullah.
\newblock Convergence guarantees for {RMSP}rop and {ADAM} in non-convex
  optimization and an empirical comparison to {N}esterov acceleration.
\newblock \emph{arXiv preprint arXiv:1807.06766}, 2018.

\bibitem[Tong et~al.(2019)Tong, Liang, and Bi]{tong2019calibrating}
Qianqian Tong, Guannan Liang, and Jinbo Bi.
\newblock Calibrating the adaptive learning rate to improve convergence of
  {ADAM}.
\newblock \emph{arXiv preprint arXiv:1908.00700}, 2019.

\bibitem[Barakat and Bianchi(2020)]{barakat2020convergence}
Anas Barakat and Pascal Bianchi.
\newblock Convergence rates of a momentum algorithm with bounded adaptive step
  size for nonconvex optimization.
\newblock In \emph{Asian Conference on Machine Learning}, pages 225--240. PMLR,
  2020.

\bibitem[Chen et~al.(2019)Chen, Liu, Sun, and Hong]{chen2018convergence}
Xiangyi Chen, Sijia Liu, Ruoyu Sun, and Mingyi Hong.
\newblock On the convergence of a class of adam-type algorithms for non-convex
  optimization.
\newblock In \emph{International Conference on Learning Representations}, 2019.

\bibitem[Barakat and Bianchi(2021)]{barakat2021convergence}
Anas Barakat and Pascal Bianchi.
\newblock Convergence and dynamical behavior of the {ADAM} algorithm for
  nonconvex stochastic optimization.
\newblock \emph{SIAM Journal on Optimization}, 31\penalty0 (1):\penalty0
  244--274, 2021.

\bibitem[Su et~al.(2014)Su, Boyd, and Candes]{su2014differential}
Weijie Su, Stephen Boyd, and Emmanuel Candes.
\newblock A differential equation for modeling {N}esterov’s accelerated
  gradient method: Theory and insights.
\newblock \emph{Advances in neural information processing systems},
  27:\penalty0 2510--2518, 2014.

\bibitem[Wibisono et~al.(2016)Wibisono, Wilson, and
  Jordan]{wibisono2016variational}
Andre Wibisono, Ashia~C Wilson, and Michael~I Jordan.
\newblock A variational perspective on accelerated methods in optimization.
\newblock \emph{proceedings of the National Academy of Sciences}, 113\penalty0
  (47):\penalty0 E7351--E7358, 2016.

\bibitem[Fran{\c{c}}a et~al.(2021)Fran{\c{c}}a, Robinson, and
  Vidal]{francca2021gradient}
Guilherme Fran{\c{c}}a, Daniel~P Robinson, and Ren{\'e} Vidal.
\newblock Gradient flows and proximal splitting methods: A unified view on
  accelerated and stochastic optimization.
\newblock \emph{Physical Review E}, 103\penalty0 (5):\penalty0 053304, 2021.

\bibitem[Shi et~al.(2021)Shi, Du, Jordan, and Su]{shi2021understanding}
Bin Shi, Simon~S Du, Michael~I Jordan, and Weijie~J Su.
\newblock Understanding the acceleration phenomenon via high-resolution
  differential equations.
\newblock \emph{Mathematical Programming}, pages 1--70, 2021.

\bibitem[Vaquero et~al.(2020)Vaquero, Mestres, and
  Cort{\'e}s]{vaquero2020resource}
Miguel Vaquero, Pol Mestres, and Jorge Cort{\'e}s.
\newblock Resource-aware discretization of accelerated optimization flows.
\newblock \emph{arXiv preprint arXiv:2009.09135}, 2020.

\bibitem[Khalil(2002)]{khalil2015nonlinear}
Hassan~K Khalil.
\newblock \emph{Nonlinear systems}.
\newblock Pearson New York, 2002.

\bibitem[Hespanha(2018)]{hespanha2018linear}
Joao~P Hespanha.
\newblock Linear systems theory.
\newblock In \emph{Linear Systems Theory}. Princeton university press, 2018.

\bibitem[Barbalat(1959)]{barbalat1959systemes}
I~Barbalat.
\newblock Systemes d’{\'e}quations diff{\'e}rentielles d’oscillations non
  lin{\'e}aires.
\newblock \emph{Rev. Math. Pures Appl}, 4\penalty0 (2):\penalty0 267--270,
  1959.

\bibitem[Mohammadi et~al.(2022)Mohammadi, Samuelson, and
  Jovanovic]{mohammadi2022transient}
Hesameddin Mohammadi, Samantha Samuelson, and Mihailo~R Jovanovic.
\newblock Transient growth of accelerated optimization algorithms.
\newblock \emph{IEEE Transactions on Automatic Control}, 2022.

\bibitem[Bottou et~al.(2018)Bottou, Curtis, and
  Nocedal]{bottou2018optimization}
L{\'e}on Bottou, Frank~E Curtis, and Jorge Nocedal.
\newblock Optimization methods for large-scale machine learning.
\newblock \emph{Siam Review}, 60\penalty0 (2):\penalty0 223--311, 2018.

\bibitem[Krizhevsky et~al.(2009)Krizhevsky, Hinton,
  et~al.]{krizhevsky2009learning}
Alex Krizhevsky, Geoffrey Hinton, et~al.
\newblock Learning multiple layers of features from tiny images.
\newblock 2009.

\bibitem[Marcus et~al.(1993)Marcus, Santorini, and
  Marcinkiewicz]{marcus1993building}
Mitchell Marcus, Beatrice Santorini, and Mary~Ann Marcinkiewicz.
\newblock Building a large annotated corpus of {E}nglish: The {P}enn
  {T}reebank.
\newblock 1993.

\bibitem[He et~al.(2016)He, Zhang, Ren, and Sun]{he2016deep}
Kaiming He, Xiangyu Zhang, Shaoqing Ren, and Jian Sun.
\newblock Deep residual learning for image recognition.
\newblock In \emph{Proceedings of the IEEE conference on computer vision and
  pattern recognition}, pages 770--778, 2016.

\bibitem[Simonyan and Zisserman(2014)]{simonyan2014very}
Karen Simonyan and Andrew Zisserman.
\newblock Very deep convolutional networks for large-scale image recognition.
\newblock \emph{arXiv preprint arXiv:1409.1556}, 2014.

\bibitem[Ma et~al.(2015)Ma, Tao, Wang, Yu, and Wang]{ma2015long}
Xiaolei Ma, Zhimin Tao, Yinhai Wang, Haiyang Yu, and Yunpeng Wang.
\newblock Long short-term memory neural network for traffic speed prediction
  using remote microwave sensor data.
\newblock \emph{Transportation Research Part C: Emerging Technologies},
  54:\penalty0 187--197, 2015.

\bibitem[Ada(2020)]{AdaBelief}
\url{https://github.com/juntang-zhuang/Adabelief-Optimizer.git}, 2020.
\newblock Accessed: 19-September-2021.

\bibitem[Abu-Mostafa et~al.(2012)Abu-Mostafa, Magdon-Ismail, and
  Lin]{abu2012learning}
Yaser~S Abu-Mostafa, Malik Magdon-Ismail, and Hsuan-Tien Lin.
\newblock \emph{Learning from data}, volume~4.
\newblock AMLBook New York, NY, USA:, 2012.

\bibitem[Jurafsky and Martin(2021)]{jurafskyspeech}
Daniel Jurafsky and James~H Martin.
\newblock Speech and language processing: An introduction to natural language
  processing, computational linguistics, and speech recognition.
\newblock \url{https://web.stanford.edu/~jurafsky/slp3/ed3book_sep212021.pdf},
  2021.
\newblock [Third edition draft; 21-September-2021].

\end{thebibliography}

\newpage
\appendix
\section{Supplemental material}

\subsection{Proof of Proposition~\ref{prop:gadagrad}}
\label{prf:prop}

\begin{proof}
The time-derivative of $f$ along the trajectories $x(t)$ of~\eqref{eqn:x_g} is $\Dot{f}(x(t)) = (\nabla f(x(t))^T \Dot{x}(t) = \sum_{i=1}^d \nabla_i f(x(t)) \Dot{x}_{i}(t)$.
Substituting~\eqref{eqn:x_g} yields, $\Dot{f}(x(t)) =  - \sum_{i=1}^d \dfrac{\norm{\nabla_i f(x(t))}^2}{\left(x_{ci}(t)\right)^c}$.
Further utilizing~\eqref{eqn:x_g} we get,
\begin{align}
    \Dot{f}(x(t)) & =  - \sum_{i=1}^d \dfrac{\Dot{x}_{ci}(t)}{\left(x_{ci}(t)\right)^c}. \label{eqn:fdot}
\end{align}
Integrating both sides above with respect to (w.r.t) time from $0$ to $t$, we get
\begin{align}
   & f(x(t))-f(x(0)) = - \sum_{i=1}^d \int_0^t \dfrac{\Dot{x}_{ci}(s)}{\left(x_{ci}(s)\right)^c} ds. \label{eqn:f_int}
\end{align}
Since $c < 1$, upon evaluating the integral we have
\begin{align}
    f(x(t)) = f(x(0)) + \sum_{i=1}^d \dfrac{\left(x_{ci}(0)\right)^{1-c} - \left(x_{ci}(t)\right)^{1-c}}{1-c}. \label{eqn:f_eval}
\end{align}
Integrating both sides of~\eqref{eqn:x_g} w.r.t time from $0$ to $t$, we have
\begin{align*}
    x_{ci}(t) = x_{ci}(0) + \int_0^t \norm{\nabla_i f(x(s))}^2 ds, ~ i \in \{1,\ldots,d\}.
\end{align*}
Substituting the above equation in~\eqref{eqn:f_eval} proves~\eqref{eqn:ft}.

Since $x_{ci}(0) > 0$, we have $x_{ci}(t) > 0$.
The above equation implies that $x_{ci}(t)$ is non-decreasing w.r.t $t$, which combined with~\eqref{eqn:ft} and $c\in (0,1)$ implies that $f(x(t))$ is non-increasing. From Assumption~\ref{assump_1}, $f$ is bounded below. Thus, $\lim_{t \to \infty} f(x(t))$ is finite. From~\eqref{eqn:f_eval} then it follows that, $\lim_{t \to \infty} x_{c}(t)$ is finite. Thus, the above equation implies that $\norm{\nabla f(x(t))}^2$ is integrable. Hence, $\nabla f(x(t))$ is bounded. 

Since $\nabla f(x(t))$ is bounded and $x_{ci}(t) > 0$, from~\eqref{eqn:x_g} we have that $\Dot{x}(t)$ is bounded.
Now, the time-derivative of $\norm{\nabla f}^2$ along the trajectories $x(t)$ is given by $\dfrac{d}{dt} \norm{\nabla f(x(t))}^2 = 2\nabla f(x(t))^T \nabla^2 f(x(t)) \Dot{x}(t)$.
We have shown that $\nabla f(x(t))$ and $\Dot{x}(t)$ are bounded. From Assumption~\ref{assump_2}, we have all the entries in $\nabla^2 f(x(t))$ are bounded. Then, from the above equation we have $\dfrac{d}{dt} \norm{\nabla f(x(t))}^2$ is bounded. Thus, $\norm{\nabla f(x(t))}^2$ is uniformly continuous.

We have shown that $\norm{\nabla f(x(t))}^2$ is integrable and $\norm{\nabla f(x(t))}^2$ is uniformly continuous. From Barbalat's lemma~\cite{barbalat1959systemes} it follows that $\lim_{t \to \infty} \norm{\nabla f(x(t))}^2 = 0$. 
\end{proof}

\subsection{Proof of Theorem~\ref{thm:gen}}
\label{prf:gen}

\begin{proof}
The time-derivative of $f$ along the trajectories $x(t)$  of~\eqref{eqn:xm_evol} is given by
\begin{align}
    & \Dot{f}(x(t)) = \sum_{i=1}^d \nabla_i f(x(t)) \Dot{x}_{i}(t) = - \sum_{i=1}^d \dfrac{1}{\alpha_g(t)} \dfrac{\lambda_7 \nabla_i f(x(t)) \mu_i(t) + \lambda_8 \norm{\nabla_i f(x(t))}^2}{\nu_i(t)^c}. \label{eqn:fdot_2}
\end{align}
Upon multiplying both sides above with $\alpha_g(t)$, followed by integrating both sides w.r.t. time from $0$ to $t$ and substituting from~\eqref{eqn:mu_evol} we have
\begin{align}
    \int_0^t \alpha_g(s)\Dot{f}(x(s))ds = & - \sum_{i=1}^d \int_0^t \dfrac{\lambda_7\mu_i(s)\Dot{\mu}_i(s)}{\lambda_2 \nu_i(s)^c}ds \nonumber\\
    & - \sum_{i=1}^d \int_0^t \dfrac{\lambda_7\lambda_1\mu_i(s)^2}{\lambda_2\nu_i(s)^c}ds  -  \sum_{i=1}^d \int_0^t \dfrac{\lambda_8\norm{\nabla_i f(x(s))}^2}{\nu_i(s)^c}ds. \label{eqn:total_int}
\end{align}
Integrating by parts we have the first term on R.H.S. as
\begin{align*}
    & \int_0^t \dfrac{\mu_i(s)\Dot{\mu}_i(s)}{\nu_i(s)^c}ds = \left[\dfrac{\mu_i(s)^2}{2\nu_i(s)^c}\right]_0^t + \dfrac{c}{2} \int_0^t \mu_i(s)^2 \nu_i(s)^{-c-1} \Dot{\nu}_i(s) ds.
\end{align*}
Upon substituting above from~\eqref{eqn:nu_evol}, and using that $\mu(0)=0_d$,
\begin{align*}
    \int_0^t \dfrac{\mu_i(s)\Dot{\mu}_i(s)}{\nu_i(s)^c}ds
    = & \dfrac{\mu_i(t)^2}{2\nu_i(t)^c} + \dfrac{c\lambda_4}{2} \int_0^t \mu_i(s)^2 \nu_i(s)^{-c-1} \zeta_i(s) ds \\
    & - \dfrac{c\lambda_5}{2}\int_0^t \mu_i(s)^2 \nu_i(s)^{-c} ds + \dfrac{c\lambda_6}{2}\int_0^t \mu_i(s)^2 \nu_i(s)^{-c-1} \psi(\nabla_i f(x(s)), \mu_i(s)) ds.
\end{align*}
Upon substituting above in~\eqref{eqn:total_int} we obtain that
\begin{align}
    \int_0^t \alpha_g(s)\Dot{f}(x(s))ds
    = & - \sum_{i=1}^d \dfrac{\lambda_7}{2\lambda_2}\mu_i(t)^2\nu_i(t)^{-c} - \sum_{i=1}^d \dfrac{c\lambda_7\lambda_4}{2\lambda_2} \int_0^t \mu_i(s)^2 \nu_i(s)^{-c-1} \zeta_i(s) ds \nonumber \\
    & - \sum_{i=1}^d \dfrac{\lambda_7}{\lambda_2} \left(\lambda_1-\dfrac{c\lambda_5}{2}\right)\int_0^t \mu_i(s)^2 \nu_i(s)^{-c} ds \nonumber \\
    & - \sum_{i=1}^d \dfrac{c\lambda_7\lambda_6}{2\lambda_2}\int_0^t \mu_i(s)^2 \nu_i(s)^{-c-1} \psi(\nabla_i f(x(s)), \mu_i(s)) ds \nonumber \\
    & - \sum_{i=1}^d \int_0^t \lambda_8\norm{\nabla_i f(x(s))}^2 \nu_i(s)^{-c}ds. \label{eqn:int_expand}
\end{align}


We consider the possible cases $\lambda_7 = 0$ and $\lambda_7 > 0$. First we consider $\lambda_7 > 0$.
We define, $\gamma_1 = 1-\lambda_2$ and $\gamma_2 = 1-\lambda_6$.
Upon differentiating both sides of~\eqref{eqn:alpha_g} w.r.t $t$ we get $\Dot{\alpha_g}(t) = \dfrac{c\gamma_2^{t+1}(1-\gamma_1^{t+1})\log{\gamma_2} - \gamma_1^{t+1}(1-\gamma_2^{t+1})\log{\gamma_1}}{(1-\gamma_2^{t+1})^{1.5}}$.
So we have
\begin{align}
    \Dot{\alpha_g}(t) < 0 \iff \left(\dfrac{\gamma_2}{\gamma_1}\right)^{t+1} \dfrac{1-\gamma_1^{t+1}}{1-\gamma_2^{t+1}} > \dfrac{1}{c}\dfrac{\log{\gamma_1}}{\log{\gamma_2}}. \label{eqn:cond}
\end{align}
From the condition $\lambda_2 > \lambda_6$, we have $1 > \gamma_2 > \gamma_1 > 0$. Then, $\left(\dfrac{\gamma_2}{\gamma_1}\right)^{t+1}$ and $\dfrac{1-\gamma_1^{t+1}}{1-\gamma_2^{t+1}}$ are, respectively, increasing and decreasing functions of $t$. Since $1 > \gamma_2 > \gamma_1 > 0$, we have $\lim_{t \to \infty} \left(\dfrac{\gamma_2}{\gamma_1}\right)^t \to \infty$ and $\lim_{t \to \infty} \dfrac{1-\gamma_1^t}{1-\gamma_2^t} = 1$. Thus, $\left(\dfrac{\gamma_2}{\gamma_1}\right)^t \dfrac{1-\gamma_1^t}{1-\gamma_2^t}$ is increasing in $t \geq T'$ for some $T'<\infty$. Then, there exists $T \in [T',\infty)$ such that~\eqref{eqn:cond} holds for all $t \geq T$. 
Integrating by parts we rewrite the L.H.S. in~\eqref{eqn:int_expand},
\begin{align*}
    & \int_0^t \alpha_g(s)\Dot{f}(x(s))ds = \left[\alpha_g(s)f(x(s))\right]_0^t - \int_0^t \Dot{\alpha_g}(s)f(x(s))ds.
\end{align*}
Upon substituting from above in~\eqref{eqn:int_expand}, for $t \geq T$,
\begin{align}
    & \alpha_g(t)f(x(t)) + \sum_{i=1}^d \dfrac{\lambda_7}{2\lambda_2}\mu_i(t)^2\nu_i(t)^{-c} \nonumber \\
    = & \alpha_g(0)f(x(0)) + \int_0^{T} \Dot{\alpha_g}(s)f(x(s))ds + \int_{T}^t \Dot{\alpha_g}(s)f(x(s))ds \nonumber \\
    & - \sum_{i=1}^d \dfrac{c\lambda_7\lambda_4}{2\lambda_2} \int_0^t \mu_i(s)^2 \nu_i(s)^{-c-1} \zeta_i(s) ds - \sum_{i=1}^d \dfrac{\lambda_7}{\lambda_2} \left(\lambda_1-\dfrac{c\lambda_5}{2}\right)\int_0^t \mu_i(s)^2 \nu_i(s)^{-c} ds \nonumber \\
    & - \sum_{i=1}^d \dfrac{c\lambda_7\lambda_6}{2\lambda_2}\int_0^t \mu_i(s)^2 \nu_i(s)^{-c-1} \psi(\nabla_i f(x(s)), \mu_i(s)) ds - \sum_{i=1}^d \int_0^t \lambda_8\norm{\nabla_i f(x(s))}^2 \nu_i(s)^{-c}ds. \label{eqn:alphaV}
\end{align}

For each $i \in \{1,\ldots,d\}$, consider the state-space system described by~\eqref{eqn:z_evol}-\eqref{eqn:nu_evol}, with the state vector $[\zeta_i,\nu_i]^T \in \R^2$. From~\eqref{eqn:lambda_cond}, $\lambda_4 \geq 0$. Here, we consider the possible cases: $\lambda_4 > 0$ and $\lambda_4 = 0$.
\begin{itemize}
    \item $\lambda_4 > 0$ : Define the state matrix $A = \begin{bmatrix} -\lambda_3 & \lambda_3 \\ \lambda_4 & -\lambda_5 \end{bmatrix}$. Since $\zeta(0) = 0_d$, the states are given by the solution
\begin{align*}
    \begin{bmatrix} \zeta_i(t) \\ \nu_i(t) \end{bmatrix} & = e^{A t} \begin{bmatrix} 0 \\ \nu_i(0)\end{bmatrix} + \int_0^t e^{A (t-s)} \begin{bmatrix} 0 \\ \lambda_6 \end{bmatrix} \psi(\nabla_i f(x(s)), \mu_i(s)) d s.
\end{align*}
Upon computing the state-transition matrix $\phi(t) = e^{A t}$ and substituting above, we obtain that
\begin{align}
    \zeta_i(t) & = \phi_{12}(t-1) \nu_i(1) + \lambda_2 \int_1^t \phi_{12}(t-s) \psi(\nabla_i f(x(s)), \mu_i(s)) d s, \label{eqn:zt} \\
    \nu_i(t) & = \phi_{22}(t-1) \nu_i(1) + \lambda_2 \int_1^t \phi_{22}(t-s) \psi(\nabla_i f(x(s)), \mu_i(s)) d s, \label{eqn:nut}
\end{align}
where $\phi_{12}(t) = \lambda_3 e^{-\frac{\lambda_3+\lambda_5}{2}t} \frac{e^{\frac{p}{2}t}-e^{-\frac{p}{2}t}}{p}$, $\phi_{22}(t) = e^{-\frac{\lambda_3+\lambda_5}{2}t} \frac{e^{\frac{p}{2}t}(p-\lambda_5+\lambda_3) + e^{-\frac{p}{2}t}(p+\lambda_5-\lambda_3)}{2p}$, and $p = \sqrt{(\lambda_3-\lambda_5)^2 + 4\lambda_3\lambda_4}$. Since $\lambda_3,\lambda_6 > 0$ (ref.~\eqref{eqn:lambda_cond}), $\nu_i(0) > 0$, $\psi \geq 0$, and $p > 0$, from~\eqref{eqn:zt} we have $\zeta_i(t) > 0$. Since $\lambda_3, \lambda_4 > 0$, we have $p = \sqrt{(\lambda_3-\lambda_5)^2 + 4\lambda_3\lambda_4} > |\lambda_5-\lambda_3|$, which implies that $\phi_{22}(t) > 0$. From above and $\nu_i(0) > 0$, we have $\nu_i(t) > 0$.
    \item $\lambda_4 = 0$ : From~\eqref{eqn:nu_evol}, $\Dot{\nu}_i(t) = - \lambda_5 \nu_i(t) + \lambda_6 \psi(\nabla_i f(x(t)), \mu_i(t))$. The solution of this ODE is $\nu_i(t) = e^{-\lambda_5 t} \nu_i(0) + \lambda_6 \int_0^t e^{-\lambda_5 (t-s)} \psi(\nabla_i f(x(s)), \mu_i(s)) d s$, which implies that $\nu_i(t) > 0$.
\end{itemize}

Since $\zeta_i(t) > 0$ and $\psi \geq 0$, due to~\eqref{eqn:lambda_cond} and~\eqref{eqn:cond}, the R.H.S. in~\eqref{eqn:alphaV} is decreasing in $t \geq T$. Then, the L.H.S. in~\eqref{eqn:alphaV} is also decreasing in $t \geq T$. Since $\mu_i(t)$ and $\nu_i(t)$ are continuous and $\nu_i(t) > 0$, $\dfrac{\lambda_7}{2\lambda_2}\mu_i(t)^2\nu_i(t)^{-c}$ is continuous. Also, $\alpha_g(t)f(x(t))$ is continuous. Thus, considering the compact interval $[0,T]$, $\alpha_g(T)f(x(T)) + \sum_{i=1}^d \dfrac{\lambda_7}{2\lambda_2}\mu_i(T)^2\nu_i(T)^{-c} =: M_T$ is finite. Since the L.H.S. in~\eqref{eqn:alphaV} is decreasing in $t \geq T$, we have the L.H.S. in~\eqref{eqn:alphaV} bounded above by $M_T$ for all $t \geq T$. 

From the R.H.S. of~\eqref{eqn:alphaV} then we have that $\mu_i(t)^2 \nu_i(t)^{-c-1} \psi(\nabla_i f(x(t)), \mu_i(t))$ and $\mu_i(t)^2 \nu_i(t)^{-c}$ are integrable w.r.t. $t$ and bounded. Moreover, if $\lambda_8 > 0$, we also have $\norm{\nabla_i f(x(t))}^2 \nu_i(t)^{-c}$ integrable and bounded. It implies that, $\psi(\nabla_i f(x(t)), \mu_i(t))$ is bounded unless $\mu_i(t) = 0$ or $\nu_i(t) = \infty$. To show that $\psi(\nabla_i f(x(t)), \mu_i(t))$ is bounded even if $\mu_i(t) = 0$ or $\nu_i(t) = \infty$, we consider the possible cases: $\lambda_8 = 0$ and $\lambda_8 > 0$.
\begin{itemize}
    \item $\lambda_8 = 0$ : From~\eqref{eqn:xm_evol}, either of the conditions $\mu_i(t) = 0$ and $\nu_i(t) = \infty$ implies that $\dot{x}_i(t) = 0$ and, hence, $\frac{d}{dt}\nabla_i f(x(t)) = 0$. Due to continuity of $\nabla_i f$, we then have $\nabla_i f(x(t))$ is bounded. Upon solving~\eqref{eqn:mu_evol}, $\mu_i(t) = \lambda_2 \int_0^t e^{-\lambda_1 (t-s)} \nabla_i f(x(s) ds$, and hence, $\mu_i(t)$ is bounded. From Assumption~\ref{assump_a}, $\nabla_i f(x(t))$ and $\mu_i(t)$ being bounded implies that $\psi(\nabla_i f(x(t)), \mu_i(t))$ is bounded.  
    \item $\lambda_8 > 0$ : In this case, $\norm{\nabla_i f(x(t))}^2 \nu_i(t)^{-c}$ is bounded. Thus, $\nabla_i f(x(t))$ is bounded unless $\nu_i(t) = \infty$, in which case $\dot{x}_i(t) = 0$. Following the argument in the previous case, $\psi(\nabla_i f(x(t)), \mu_i(t))$ is bounded. 
\end{itemize}
Thus, $\psi(\nabla_i f(x(t)), \mu_i(t))$ is bounded, and therefore, Assumption~\ref{assump_b} implies that $\nabla_i f(x(t))$ is bounded.

To show that $\nu_i(t)$ is bounded, we consider the possible cases: $\lambda_4 > 0$ and $\lambda_4 = 0$.
\begin{itemize}
    \item $\lambda_4 > 0$ : We can rewrite $\phi_{22}$ as $\phi_{22}(t) = \frac{e^{-\frac{\lambda_3+\lambda_5-p}{2}t}(p-\lambda_5+\lambda_3) + e^{-\frac{\lambda_3+\lambda_5+p}{2}t}(p+\lambda_5-\lambda_3)}{2p}$. Since $p = \sqrt{(\lambda_3-\lambda_5)^2 + 4\lambda_3\lambda_4} \leq \lambda_3+\lambda_5$ for $\lambda_5 \geq \lambda_4$ (see~\eqref{eqn:lambda_cond}) and $\psi(\nabla_i f(x(t)), \mu_i(t))$ is bounded, it follows from~\eqref{eqn:nut} that $\nu_i(t)$ is bounded.
    \item $\lambda_4 = 0$ : In this case, $\nu_i(t) = e^{-\lambda_5 t} \nu_i(0) + \lambda_6 \int_0^t e^{-\lambda_5 (t-s)} \psi(\nabla_i f(x(s)), \mu_i(s)) ds$. The above implies that $\nu_i(t)$ is bounded as $\psi(\nabla_i f(x(t)), \mu_i(t))$ is bounded.
\end{itemize}
Earlier, we have shown that $\mu_i(t)$ is bounded and $\nu_i(t) > 0$. From~\eqref{eqn:xm_evol} then we have, $\Dot{x}_i(t)$ is bounded. From~\eqref{eqn:mu_evol}, $\mu_i(t) = 0$ implies that $\dot{\mu}_i(t) = \lambda_1 \nabla_i f(x(t))$. Thus, $\mu_i(t)$ can be zero only at isolated points $t$. Otherwise, for some $h>0$ there exists an interval $(t-h,t+h)$ such that $\mu_i(s) = 0$ for all $s \in (t-h,t+h)$. In that case, $\dot{\mu}_i(s) = 0$ for all $s \in (t-h,t+h)$. Since $\dot{\mu}_i(s) = \lambda_1 \nabla_i f(x(s))$ for all $s \in (t-h,t+h)$, we then have $\nabla_i f(x(s)) = 0$ for all $s \in (t-h,t+h)$, which proves the theorem.

We have shown above that $\mu_i(t) = 0$ only at isolated points and $\nu_i(t)$ is bounded. So, $\dfrac{1}{\mu_i(t)^2 \nu_i(t)^{-c}}$ is bounded except at isolated points.
Since $\mu_i(t)^2 \nu_i(t)^{-c}$ is integrable and $\dfrac{1}{\mu_i(t)^2 \nu_i(t)^{-c}}$ is bounded except at isolated points, we have $\dfrac{1}{\mu_i(t)^2 \nu_i(t)^{-c}}$ is integrable. Since $\nu_i(t)$ is bounded and $\dfrac{1}{\mu_i(t)^2 \nu_i(t)^{-c}}$ is integrable, we have $\dfrac{1}{\mu_i(t)^2 \nu_i(t)^{-c-1}}$ integrable.
Now, we apply Cauchy-Schwartz inequality on the functions $\mu_i(t) \nu_i(t)^{(-c-1)/2} \psi(\nabla_i f(x(t)), \mu_i(t))^{0.5}$ and $\dfrac{1}{\mu_i(t) \nu_i(t)^{(-c-1)/2}}$. Since we have $\mu_i(t)^2 \nu_i(t)^{-c-1} \psi(\nabla_i f(x(t)), \mu_i(t))$ and $\dfrac{1}{\mu_i(t)^2 \nu_i(t)^{-c-1}}$ integrable, the Cauchy-Schwartz inequality implies that $\psi(\nabla_i f(x(t)), \mu_i(t))^{0.5}$ is integrable. Since $\psi(\nabla_i f(x(t)), \mu_i(t))^{0.5}$ is bounded and integrable, it is also square-integrable. Thus, $\psi(\nabla_i f(x(t)), \mu_i(t))$ is integrable.

Now, the time-derivative of $\psi$ along the trajectory $x_i(t)$ is
\begin{align*}
    & \dfrac{d}{dt} \psi(\nabla_i f(x(t)), \mu_i(t)) = \nabla \psi(\nabla_i f(x(t)), \mu_i(t))^T 
    \begin{bmatrix}
    \left[\nabla^2 f(x(t))\right]_i \Dot{x}(t) \\ \Dot{\mu}_i(t)
    \end{bmatrix}.
\end{align*}
We have shown that $\nabla_i f(x(t))$ and $\mu_i(t)$ are bounded. Then, according to Assumption~\ref{assump_a}, $\nabla \psi(\nabla_i f(x(t)), \mu_i(t))$ is bounded. From Assumption~\ref{assump_2}, $[\nabla^2 f(x(t))]_i$ is bounded. We have shown that $\Dot{x}_i(t)$ is bounded. Since $\nabla_i f(x(t))$ and $\mu_i(t)$ are bounded,~\eqref{eqn:mu_evol} implies that $\Dot{\mu}_i(t)$ is bounded. Then, from the above equation we have $\dfrac{d}{dt} \psi(\nabla_i f(x(t)), \mu_i(t))$ is bounded. Thus, $\psi(\nabla_i f(x(t)), \mu_i(t))$ is uniformly continuous w.r.t $t$.

We have shown that, for each $i \in \{1,\ldots,d\}$, $\psi(\nabla_i f(x(t)), \mu_i(t))$ is integrable and uniformly continuous. From Barbalat's lemma~\cite{barbalat1959systemes} it follows that $\lim_{t \to \infty} \psi(\nabla_i f(x(t)), \mu_i(t)) = 0$. Then, according to Assumption~\ref{assump_c}, $\lim_{t \to \infty} \nabla_i f(x(t)) = 0$.

Next, we consider the case $\lambda_7 = 0$.  In this case, $\alpha_g(t) = 1$ (see~\eqref{eqn:alpha_g}). Also, from~\eqref{eqn:lambda_cond}, $\lambda_8 > 0$. For $\lambda_7 = 0$, the argument in the paragraph following~\eqref{eqn:alphaV} still holds, and we obtain that $\nu_i(t)>0$. Upon substituting $\lambda_7 = 0$ and $\alpha_g(t) = 1$ in~\eqref{eqn:int_expand},
\begin{align*}
    f(x(t)) = f(x(0)) - \sum_{i=1}^d \int_0^t \lambda_8\norm{\nabla_i f(x(s))}^2 \nu_i(s)^{-c}ds.
\end{align*}
Then, $f(x(t))$ is decreasing in $t$. From Assumption~\ref{assump_1}, $f$ is bounded below. Thus, $\lim_{t \to \infty} f(x(t))$ is finite. So, the above equation implies that $\norm{\nabla_i f(x(t))}^2 \nu_i(t)^{-c}$ is integrable. So, from the previous argument for the case $\lambda_8>0$, $\nabla_i f(x(t))$ is bounded. From~\eqref{eqn:xm_evol} then we have, $\Dot{x}_i(t)$ is bounded. Upon solving~\eqref{eqn:mu_evol}, $\mu_i(t) = \lambda_2 \int_0^t e^{-\lambda_1 (t-s)} \nabla_i f(x(s) ds$, and hence, $\mu_i(t)$ is bounded. Now,
\begin{align}
    & \frac{d}{dt} \norm{\nabla_i f(x(t))}^2 \nu_i(t)^{-c} \nonumber \\
    = & - c \norm{\nabla_i f(x(t))}^2 \nu_i(t)^{-c-1} \Dot{\nu}_i(t) + 2\nabla_i f(x(t)) \left[\nabla^2 f(x(t))\right]_i \Dot{x}(t) \nu_i(t)^{-c}. \label{eqn:grad_diff}
\end{align}
From Assumption~\ref{assump_2}, $[\nabla^2 f(x(t))]_i$ is bounded. From Assumption~\ref{assump_a}, $\nabla_i f(x(t))$ and $\mu_i(t)$ being bounded implies that $\psi(\nabla_i f(x(t)), \mu_i(t))$ is bounded. From the previous arguments for the cases $\lambda_4 = 0$ and $\lambda_4 > 0$, it follows that $\nu_i(t)$ and $\Dot{\nu}_i(t)$ are bounded. So,~\eqref{eqn:grad_diff} implies that $\frac{d}{dt} \norm{\nabla_i f(x(t))}^2 \nu_i(t)^{-c}$ is bounded. Then, $\norm{\nabla_i f(x(t))}^2 \nu_i(t)^{-c}$ is uniformly continuous. Again, we apply Barbalat's lemma~\cite{barbalat1959systemes} and obtain $\lim_{t \to \infty} \norm{\nabla_i f(x(t))}^2 \nu_i(t)^{-c} = 0$. Since $\nu_i(t)$ is bounded, we get $\lim_{t \to \infty} \norm{\nabla_i f(x(t))} = 0$.
The proof is complete.
\end{proof}

\subsection{Proof of Corollary~\ref{cor:gadagrad}}
\label{prf:gadagrad}

\begin{proof}
The set of equations~\eqref{eqn:x_g} is a special case of~\eqref{eqn:mu_evol}-\eqref{eqn:xm_evol} with $\nu = x_c$, $\lambda_4 = \lambda_5 = 0$, $\lambda_6 = 1$, $\lambda_7 = 0$, $\lambda_8 = 1$, and $\psi(\nabla_i f(x(t)), \mu_i(t)) = \norm{\nabla_i f(x(t))}^2$. Clearly, $\psi(\nabla_i f(x(t)), \mu_i(t)) = \norm{\nabla_i f(x(t))}^2$ satisfies Assumption~\ref{assump_3}. Thus, Theorem~\ref{thm:gen} is applicable, and the proof follows.
\end{proof}

\subsection{Proof of Corollary~\ref{cor:adam}}
\label{prf:adam}

\begin{proof}
The set of equations~\eqref{eqn:xm_adam} is a special case of~\eqref{eqn:mu_evol}-\eqref{eqn:xm_evol} with $c=0.5$, $\lambda_1 = \lambda_2 = b_1$, $\lambda_4 = 0$, $\lambda_6 = b_2$, $\lambda_7 = 1$, $\lambda_8 = 0$, and $\psi(\nabla_i f(x(t)), \mu_i(t)) = \norm{\nabla_i f(x(t))}^2$. Clearly, $\psi(\nabla_i f(x(t)), \mu_i(t)) = \norm{\nabla_i f(x(t))}^2$ satisfies Assumption~\ref{assump_3}. Thus, Theorem~\ref{thm:gen} is applicable, and the proof follows.
\end{proof}

\subsection{Proof of Corollary~\ref{cor:adabelief}}
\label{prf:adabelief}

\begin{proof}
The set of equations~\eqref{eqn:v_adabelief}-\eqref{eqn:xm_adabelief} is a special case of~\eqref{eqn:mu_evol}-\eqref{eqn:xm_evol} with $c=0.5$, $\lambda_1 = \lambda_2 = b_1$, $\lambda_4 = 0$, $\lambda_6 = b_2$, $\lambda_7 = 1$, $\lambda_8 = 0$, and $\psi(\nabla_i f(x(t)), \mu_i(t)) = \norm{\nabla_i f(x(t))-\mu_i(t)}^2$. Clearly, $\psi(\nabla_i f(x(t)), \mu_i(t)) = \norm{\nabla_i f(x(t))-\mu_i(t)}^2$ satisfies Assumption~\ref{assump_a}-\ref{assump_b}. We will show that it also satisfies Assumption~\ref{assump_c}.

Suppose that $\lim_{t \to \infty} \norm{\nabla_i f(x(t))-\mu_i(t)}^2 = 0$. Then, $\lim_{t \to \infty} \nabla_i f(x(t)) = \lim_{t \to \infty} \mu_i(t)$. Then, from~\eqref{eqn:v_adabelief}, $\lim_{t \to \infty} \Dot{\mu}_i(t) = 0$. We define the notation $\mu_i^* = \lim_{t \to \infty} \mu_i(t) = \lim_{t \to \infty} \nabla_i f(x(t))$. We claim that $\mu_i^* = 0$. Otherwise, suppose that $\mu_i^* \neq 0$. Then, there exists $T < \infty$ such that $\mu_i(t)$ and $\nabla_i f(x(t))$ have the same sign for all $t \geq T$. Then, from~\eqref{eqn:fdot_2}, $\Dot{f}(x(t)) < 0$ for all $t \geq T$. Since $f$ is bounded below, due to Assumption~\eqref{assump_1}, and decreasing for all $t \geq T$, $f(x(t))$ converges to its minimum value as $t \to \infty$. Upon differentiating both sides of~\eqref{eqn:fdot_2} w.r.t. $t$,
\begin{align}
    \Ddot{f}(x(t)) = \sum_{i=1}^d \nabla_i f(x(t)) \Ddot{x}_{i}(t) + \sum_{i=1}^d \left[\nabla^2 f(x(t))\right]_i \Dot{x}(t) \Dot{x}_i(t). \label{eqn:f_ddot}
\end{align}
Upon differentiating both sides of~\eqref{eqn:xm_adabelief} w.r.r. $t$ and substituting from~\eqref{eqn:v_adabelief},
\begin{align*}
    \Ddot{x}_i(t) = - \frac{\Dot{\mu}_i(t)}{\alpha(t) \nu_i(t)^{0.5}} + \frac{\Dot{\alpha}(t) \mu_i(t)}{\alpha(t)^2 \nu_i(t)^{0.5}} + \frac{b_2 \mu_i(t) \norm{\nabla_i f(x(t))-\mu_i(t)}^2}{2 \alpha(t) \nu_i(t)^{1.5}} - \frac{b_2 \mu_i(t)}{2 \alpha(t) \nu_i(t)^{0.5}}.
\end{align*}
From the definition of $\alpha$ in Section~\ref{sec:intro}, $\alpha(t) > 0$ and $\Dot{\alpha}(t)$ is bounded. Also, $\nu_i(t) > 0$. Since $\lim_{t \to \infty} \norm{\nabla_i f(x(t))-\mu_i(t)}^2 = \lim_{t \to \infty} \Dot{\mu}_i(t) = 0$ and $\lim_{t \to \infty} \mu_i(t) = \mu_i^*$, we have $\norm{\nabla_i f(x(t))-\mu_i(t)}^2$, $\Dot{\mu}_i(t)$, and $\mu_i(t)$ bounded. Then, $\Ddot{x}_i(t)$ is bounded. Since $\lim_{t \to \infty} \nabla_i f(x(t)) = \mu_i^*$, $\nabla_i f(x(t))$ is bounded. Hence, under Assumption~\ref{assump_2},~\eqref{eqn:f_ddot} implies that $\Ddot{f}(x(t))$ is bounded. Since we have $\lim_{t \to \infty} f(x(t))$ finite and $\Ddot{f}(x(t))$ bounded, Barbalat's lemma~\cite{barbalat1959systemes} implies that $\lim_{t \to \infty} \Dot{f}(x(t)) = 0$. But we also shown that 
$\Dot{f}(x(t)) < 0$ for all $t \geq T$. This is a contradiction. So, Assumption~\ref{assump_c} holds and the proof is complete following Theorem~\ref{thm:gen}.
\end{proof}

\subsection{Proof of Corollary~\ref{cor:AdamSSM}}
\label{prf:AdamSSM}

\begin{proof}
The set of equations~\eqref{eqn:mu_sd}-\eqref{eqn:xm_sd} is a special case of~\eqref{eqn:mu_evol}-\eqref{eqn:xm_evol} with $c=0.5$, $\lambda_1 = \lambda_2 = b_1$, $\lambda_3 = b_2$, $\lambda_4 = b_3$, $\lambda_5 = b_2 + b_3$, $\lambda_6 = b_2$, $\lambda_7 = 1$, $\lambda_8 = 0$, and $\psi(\nabla_i f(x(t)), \mu_i(t)) = \norm{\nabla_i f(x(t))}^2$. Clearly, $\psi(\nabla_i f(x(t)), \mu_i(t)) = \norm{\nabla_i f(x(t))}^2$ satisfies Assumption~\ref{assump_3}. Thus, Theorem~\ref{thm:gen} is applicable, and the proof follows.
\end{proof}

\subsection{Experimental results with additional details}
\label{sec:exp_full}

In this section, we present experimental results on benchmark machine learning problems, comparing the convergence rate and test-set accuracy~\cite{abu2012learning} of the proposed AdamSSM algorithm with several other adaptive gradient methods. These methods are AdaBelief~\cite{NEURIPS2020_d9d4f495}, AdaBound~\cite{luo2018adaptive}, Adam~\cite{kingma2014adam}, AdamW~\cite{loshchilov2018decoupled}, Fromage~\cite{bernstein2020distance}, MSVAG~\cite{balles2018dissecting}, RAdam~\cite{liu2019variance}, SGD~\cite{bottou2018optimization}, and Yogi~\cite{ZaheerRSKK18}.

We implement the proposed AdamSSM algorithm in discrete-time. Specifically, we use first-order {\em explicit} Euler discretization with a fixed sampling rate to discretize the set of ODEs~\eqref{eqn:mu_sd}-\eqref{eqn:xm_sd} which models the proposed algorithm in continuous-time. To be consistent with the format of the other algorithms, we replace the condition $\nu_i(0) > 0 \, \forall i$ with an additional parameter $\epsilon > 0$, as done in the other algorithms. The purpose of either of these conditions is the same, which is to avoid division by zero in~\eqref{eqn:xm_sd}. Additionally, we denote the learning rate parameter for updating the estimate $x(t)$ for each iteration $t = 0,1,\ldots$ in discrete-time by $\eta(t)$.
The AdamSSM algorithm in discrete-time is summarized above in Algorithm~\ref{algo_1}.

In the experiments, we consider two machine learning tasks: image classification on CIFAR-10 dataset~\cite{krizhevsky2009learning} and language modeling on Penn TreeBank (PTB) dataset~\cite{marcus1993building}. The CIFAR-10 dataset consists of $60k$ tiny colour images with $32 \times 32$ pixels in $10$ mutually exclusive classes, with $6k$ images per class. There are $50k$ training images and $10k$ test images. The PTB  dataset consists of $929k$ training words,
$73k$ validation words, and $82k$ test words.

For image classification task, we use two CNN architectures: ResNet34~\cite{he2016deep} and VGG11~\cite{simonyan2014very}. The numeral after the keyword signifies the number of weighted layers in that architecture. ResNet34 and VGG11 has approximately $d=21$ million and $d=133$ million parameters, respectively. These are the state-of-the-art architectures for image classification. ResNet, in particular, solves the famous vanishing gradient problem, where the computed gradients get truncated to zero due to repeated application of chain rule across deep layers during back-propagation and due to finite precision.

For language modeling task, we use the long short-term memory (LSTM)~\cite{ma2015long} architecture with respectively 1-layer, 2-layers, and 3-layers. LSTM is a widely used language model in different applications, including text generation and speech recognition. It is a recurrent neural network with `gates' which are neural network that learns the important information from the training data corpus. Perplexity~\cite{jurafskyspeech} is a metric for measuring performance of a language model. Technically, a language model computes the joint probability of a word sequence from product of conditional probabilities of each word. Perplexity is defined as the inverse probability of the test set, as predicted by a trained model, normalised by the number of words. Perplexity can also be interpreted as the number of words that can be encoded with the cross-entropy. Thus, lower the perplexity, more confident the model is in predicting the next word in a sequence. So, a lower perplexity is preferred. 

To conduct these experiments, we adapt the experimental setup used in the recent AdaBelief paper~\cite{NEURIPS2020_d9d4f495} and the AdaBound paper~\cite{luo2018adaptive}. The estimate $x(t)$ is initialized randomly from $\R^d$ for all the algorithms. The hyperparameters of the respective algorithms are tuned such that the individual algorithms achieves a better generalization on the test dataset. Following~\cite{NEURIPS2020_d9d4f495}, these hyperparameters are selected as described below. 

{\bf AdaBelief}: The standard parameter values $\beta_1 = 0.9$, $\beta_2 = 0.999$ are used. The parameter $\epsilon$ is set to $10^{-8}$ for image classification tasks, $10^{-16}$ for 1-layer LSTM and $10^{-12}$ for 2-layer and 3-layer LSTM. The learning rate $\eta$ is set to $10^{-2}$ for 2-layer and 3-layer LSTM, and $10^{-3}$ for all other models. These parameter values are set according to the implementation of AdaBelief in GitHub~\footnote{https://github.com/juntang-zhuang/Adabelief-Optimizer}.

{\bf AdaBound, Adam, MSVAG, RAdam, Yogi}: The parameter $\beta_1$ is selected from the set $\{0.5,0.6,0.7,0.8,0.9\}$. The learning rate $\eta$ is selected from the set $\{10^p: \, p = 1,0,-1,-2,-3\}$. Standard values are used for the other parameters. 

{\bf AdamW}: The weight decay parameter is chosen from the set $\{10^{-2}, 10^{-3}, 5 \times 10^{-4}, 10^{-4}\}$. The other parameters are selected in the same way as Adam.

{\bf Fromage, SGD}: The learning rate is selected as described above for Adam. The momentum is chosen as the default value of $0.9$.

{\bf AdamSSM}: The parameters $\beta_1 = (1-\delta b_1)$ and $\beta_2 = (1-\delta b_2)$ are similar to Adam. The standard choices for $\beta_1$ and $\beta_2$ in Adam are respectively $0.9$ and $0.999$~\cite{kingma2014adam}. With a sampling time $\delta = 0.15$, therefore we set $b_1 = 0.67$ and $b_2 = 0.0067$. The parameter $b_3$ is chosen from the set $\{\frac{c\times 10^{-3}}{\delta}: c = 1,2,3,4,5\}$. The parameter $\epsilon$ and the learning rate $\eta$ are selected in the same way as AdaBelief.

In our experiments, we have considered the {\em cross-entropy} loss function~\cite{abu2012learning}. 
To avoid overfitting, we have used $l_2$-regularization~\cite{abu2012learning} while training the architectures. Following~\cite{NEURIPS2020_d9d4f495}, the regularization hyperparameter is set to $5 \times 10^{-4}$ for the image classification tasks and $1.2 \times 10^{-6}$ for the language modeling task, for each of the aforementioned algorithms.

For the image classification tasks, the model is trained for $200$ epochs; the learning rate is multiplied by $0.1$ at epoch $150$; and a mini-batch size of $128$ is used~\cite{NEURIPS2020_d9d4f495, luo2018adaptive}. We compare the training-set and test-set accuracy of different training algorithms in Table~\ref{tab:resnet_full} and Table~\ref{tab:vgg_full}. We observe that the proposed AdamSSM algorithm has the best test-set accuracy among all the algorithms, on both the architectures ResNet34 and VGG11. Some other algorithms achieve a better training-set accuracy than the proposed method. However, the test-set accuracy of those algorithms is less than AdamSSM.

For the language modeling tasks, the model is trained for $200$ epochs; the learning rate is multiplied by $0.1$ at epoch $100$ and $145$; and a mini-batch size of $20$ is used~\cite{NEURIPS2020_d9d4f495}. We compare the training-set and test-set perplexity of different training algorithms in Table~\ref{tab:lstm1_full}-\ref{tab:lstm3_full}. Note that a lower perplexity means better accuracy. For 1-layer LSTM, only the Adam method generates lower test set perplexity than the proposed method. For 2-layer LSTM, only the AdaBelief method generates lower test set perplexity than the proposed method. For the more complex 3-layer LSTM, the proposed method achieves both the least test set and the least training set perplexity.

\begin{figure*}[htb!]
\centering
\begin{subfigure}{.5\textwidth}
  \begin{center}
  \includegraphics[width = \textwidth]{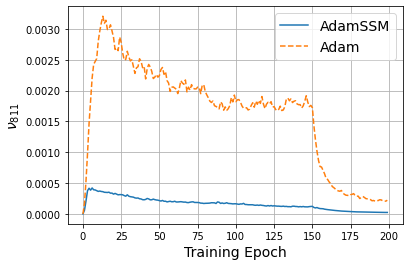}
  \caption{dimension $i=811$}
  \label{fig:nu811}
  \end{center}
\end{subfigure}%
\begin{subfigure}{.5\textwidth}
  \begin{center}
  \includegraphics[width = \textwidth]{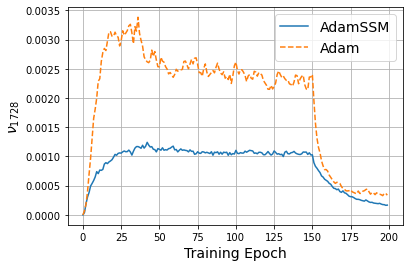}
  \caption{dimension $i=1728$}
  \label{fig:nu1728}
  \end{center}
\end{subfigure}
\caption{\it Second raw moment estimate of gradient along two different dimensions, (a) $i=811$ and (b) $i=1728$, for image classification task on CIFAR-10 dataset with VGG11 architecture trained with Adam and the proposed AdamSSM algorithms.}
\label{fig:nu}
\end{figure*}

Recall that the transfer function from the squared gradient to $\nu_i(t)$ is $\frac{b_2(s+b_2)}{s^2+(2b_2+b_3)s+b_2^2}$ for the proposed AdamSSM algorithm and $\frac{b_2}{s+b_2}$ for the Adam algorithm. So, the two poles and the single zero of this transfer function depend on the parameter values $b_2, b_3$, where $b_3$ is an additional parameter compared to Adam. By tuning the hyperparameters $b_2, b_3$ we create a signal $\nu_i(t)$ that is less impacted by the noise than the input signal of the squared gradient. The transfer functions for both the Adam and AdamSSM methods act as a low-pass filter on the squared gradient input to the output $\nu_i(t)$. So, in other words, by properly tuning the hyperparameters $b_2, b_3$, the AdamSSM algorithm attenuates frequency-specific noise in the squared gradient during the training better than Adam, which leads to improved generalization. We numerically validate our above hypothesis on the image classification task with VGG11. In Figure~\ref{fig:nu}, we have plotted the curves of $\nu_i(t)$ when the VGG11 model is trained with Adam and AdamSSM algorithms, along two different dimensions $i = 811,1728$. The output signal $\nu_i(t)$ increases initially, albeit with larger damping for AdamSSM due to the addition of an LHP pole compared to Adam. After the initial rise, $\nu_i(t)$ in the  Adam method fluctuates much more throughout the stable part of the signal. It implies that the output $\nu_i(t)$ of Adam is significantly more impacted by the noise, which supports our hypothesis that the AdamSSM algorithm supports better generalization by being more robust to noisy signals.

\begin{figure*}[htb!]
\centering
\begin{subfigure}{.5\textwidth}
  \begin{center}
  \includegraphics[width = \textwidth]{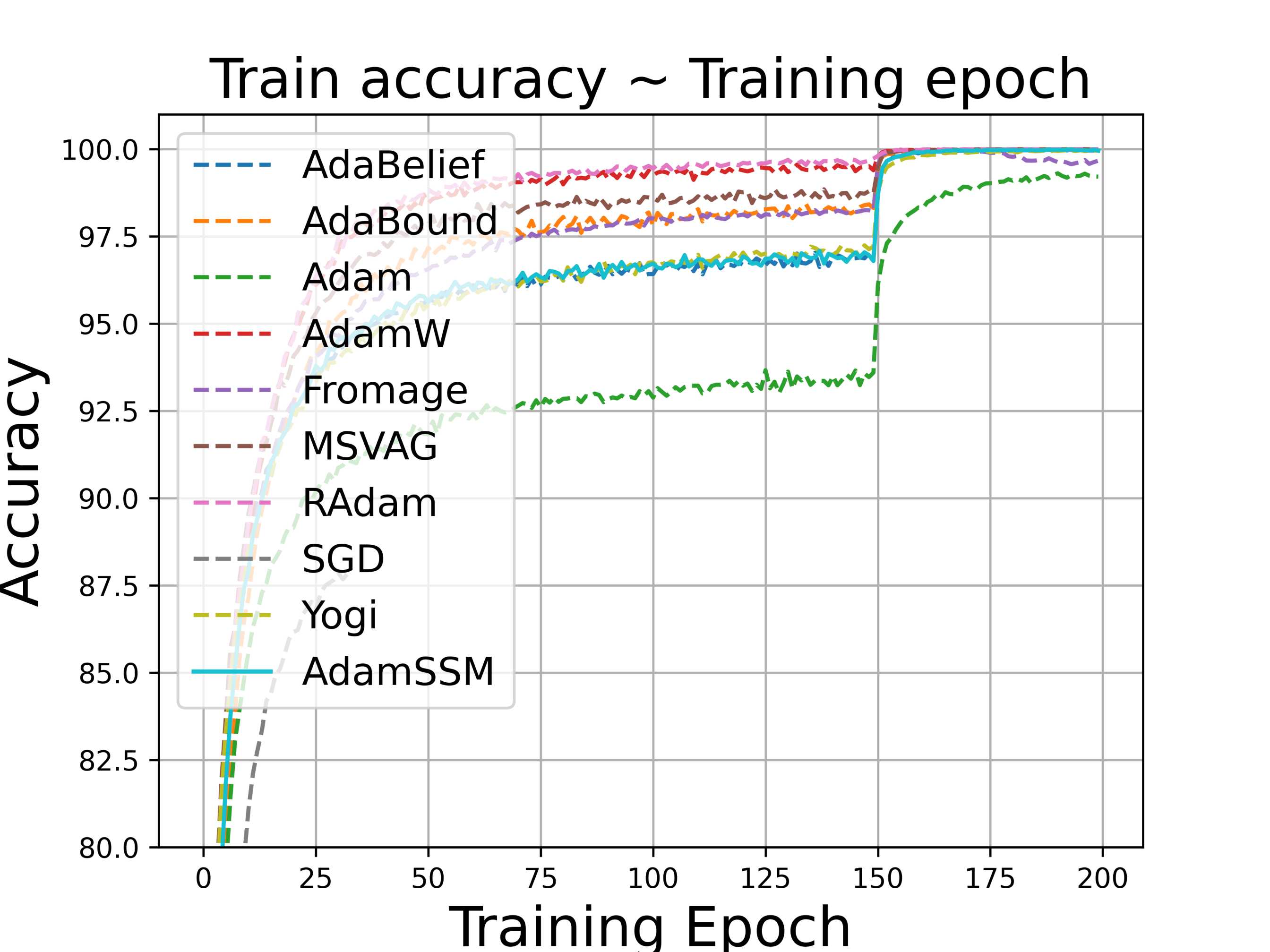}
  \caption{training data}
  \label{fig:train_resnet}
  \end{center}
\end{subfigure}%
\begin{subfigure}{.5\textwidth}
  \begin{center}
  \includegraphics[width = \textwidth]{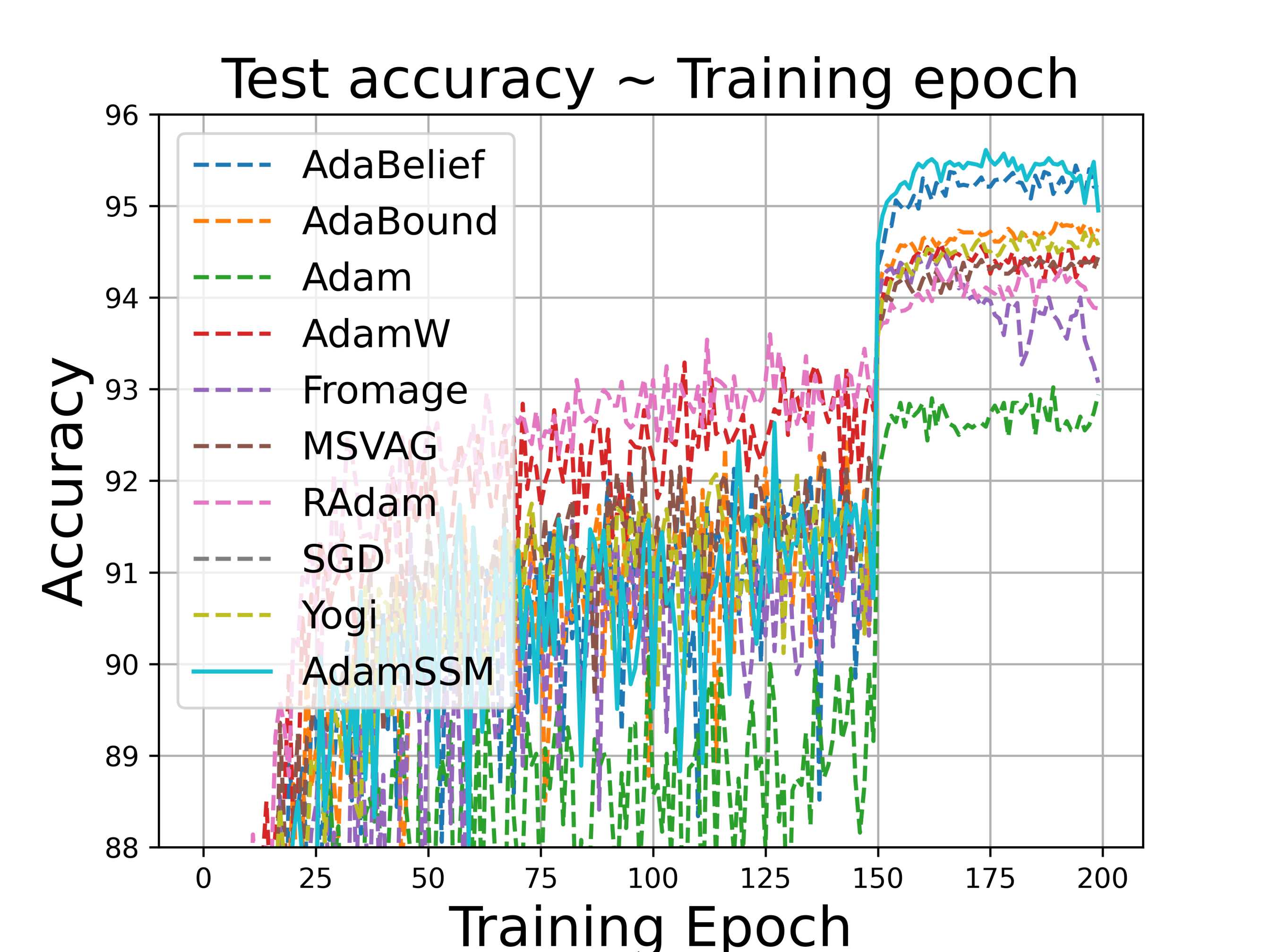}
  \caption{test data}
  \label{fig:test_resnet}
  \end{center}
\end{subfigure}
\caption{\it Accuracy for image classification task on CIFAR-10 dataset with ResNet34 architecture trained with different algorithms, for (a) training data and (b) test data.}
\label{fig:resnet}
\end{figure*}

\begin{figure*}[htb!]
\centering
\begin{subfigure}{.5\textwidth}
  \begin{center}
  \includegraphics[width = \textwidth]{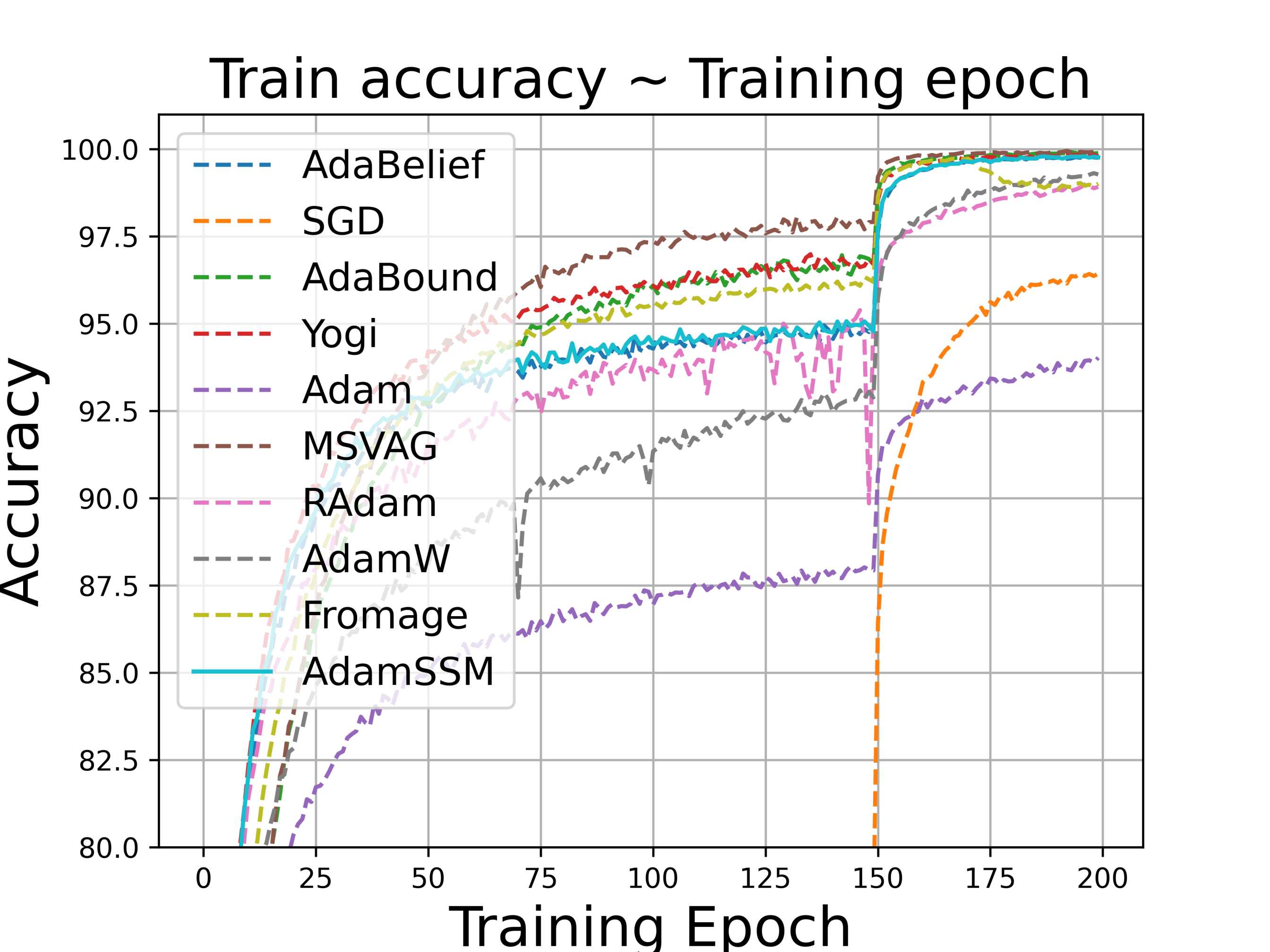}
  \caption{training data}
  \label{fig:train_vgg}
  \end{center}
\end{subfigure}%
\begin{subfigure}{.5\textwidth}
  \begin{center}
  \includegraphics[width = \textwidth]{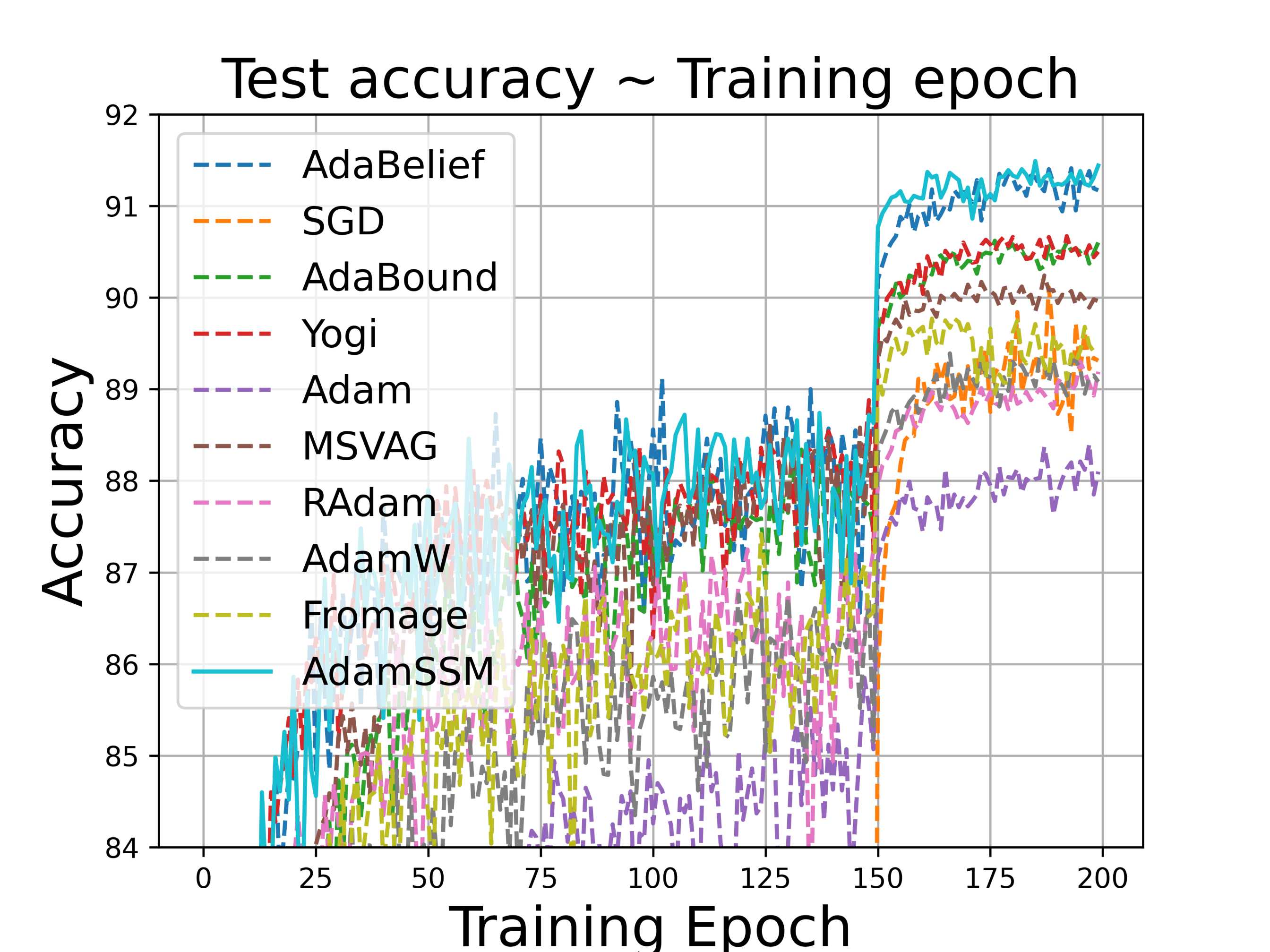}
  \caption{test data}
  \label{fig:test_vgg}
  \end{center}
\end{subfigure}
\caption{\it Accuracy for image classification task on CIFAR-10 dataset with VGG11 architecture trained with different algorithms, for (a) training data and (b) test data.}
\label{fig:vgg}
\end{figure*}

\begin{figure*}[htb!]
\centering
\begin{subfigure}{.5\textwidth}
  \begin{center}
  \includegraphics[width = \textwidth]{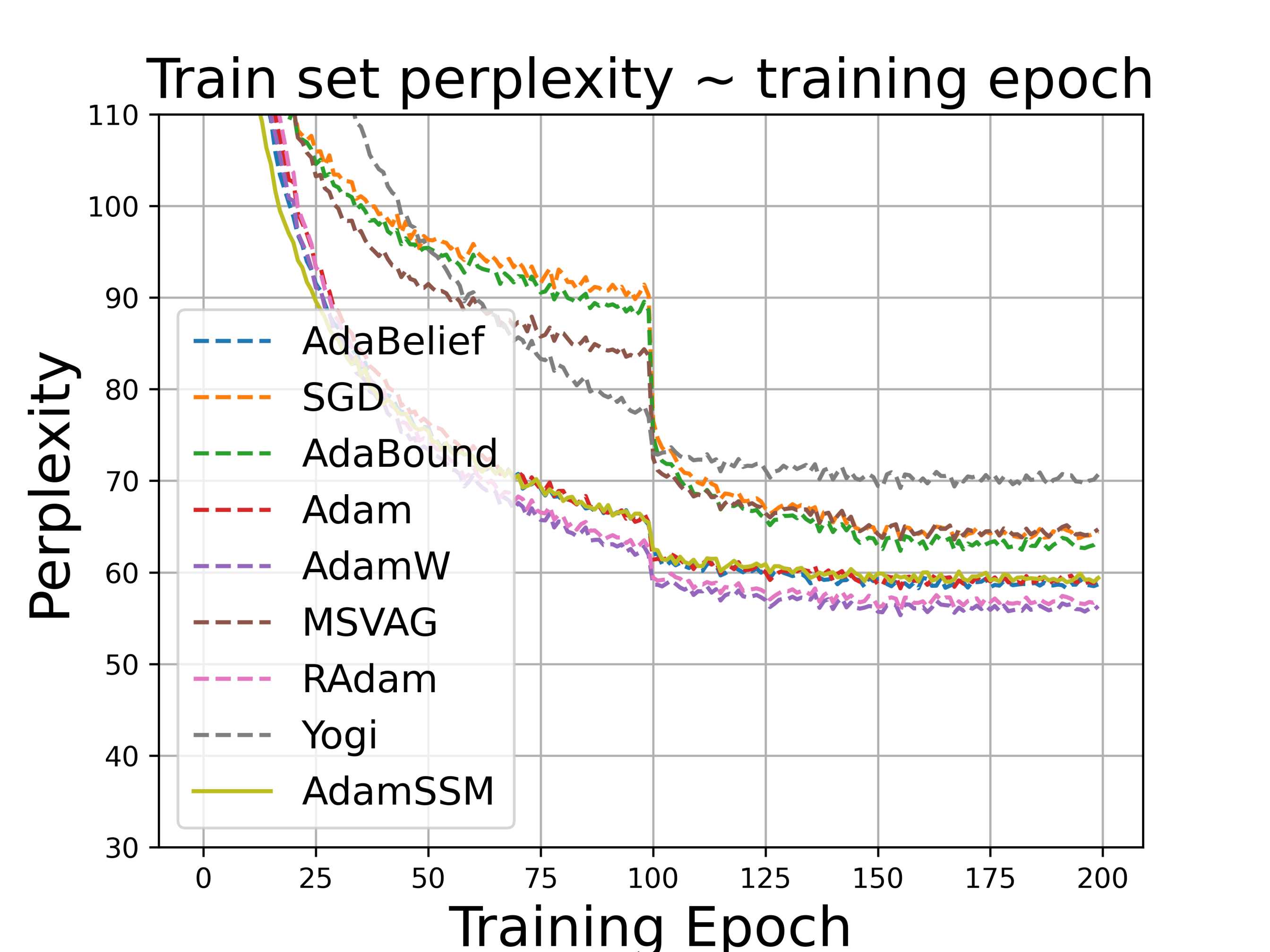}
  \caption{training data}
  \label{fig:train_1layer}
  \end{center}
\end{subfigure}%
\begin{subfigure}{.5\textwidth}
  \begin{center}
  \includegraphics[width = \textwidth]{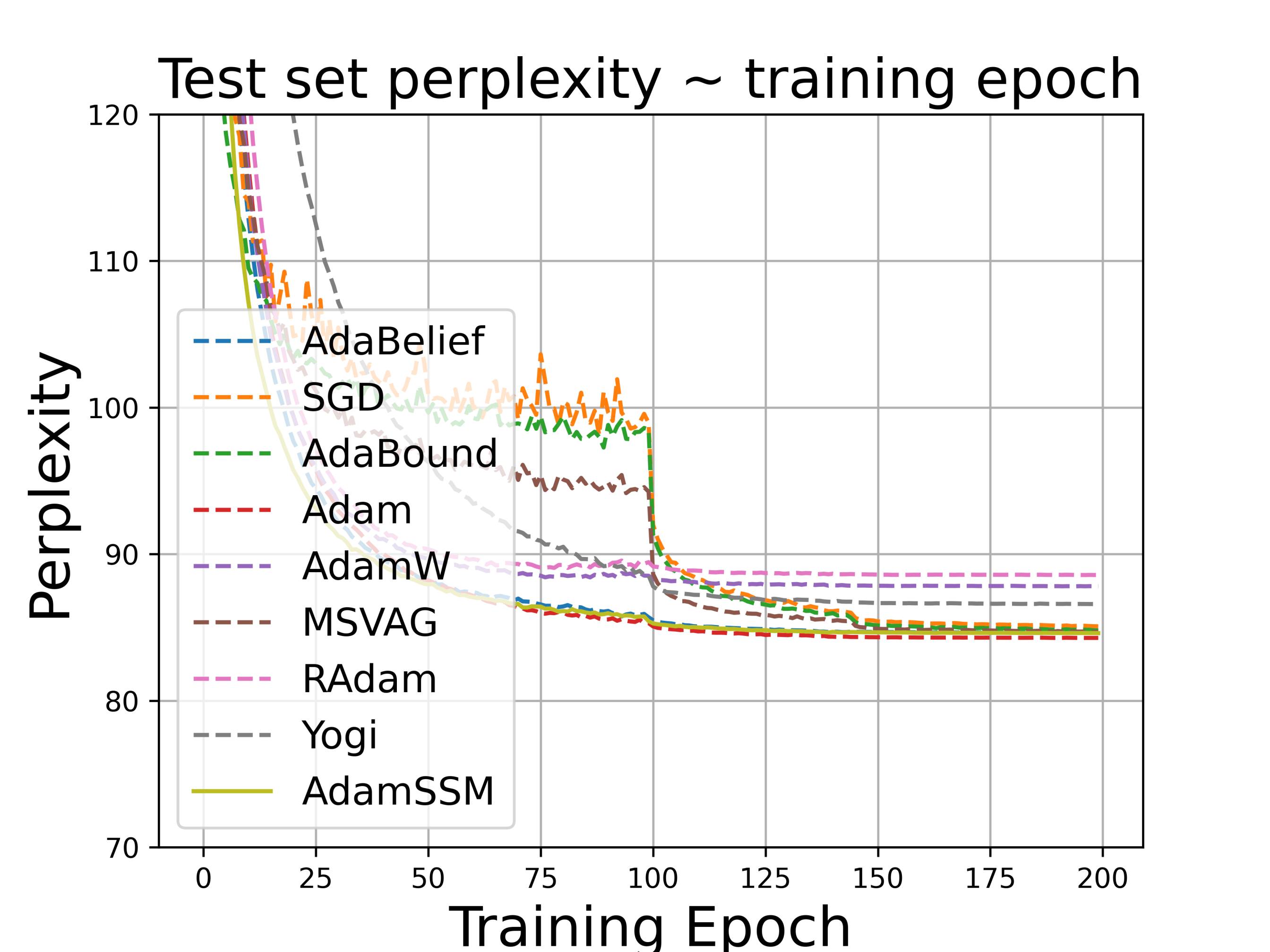}
  \caption{test data}
  \label{fig:test_1layer}
  \end{center}
\end{subfigure}
\caption{\it Accuracy for language modeling task on Penn TreeBank dataset with 1-layer LSTM architecture trained with different algorithms, for (a) training data and (b) test data.}
\label{fig:1layer}
\end{figure*}

\begin{figure*}[htb!]
\centering
\begin{subfigure}{.5\textwidth}
  \begin{center}
  \includegraphics[width = \textwidth]{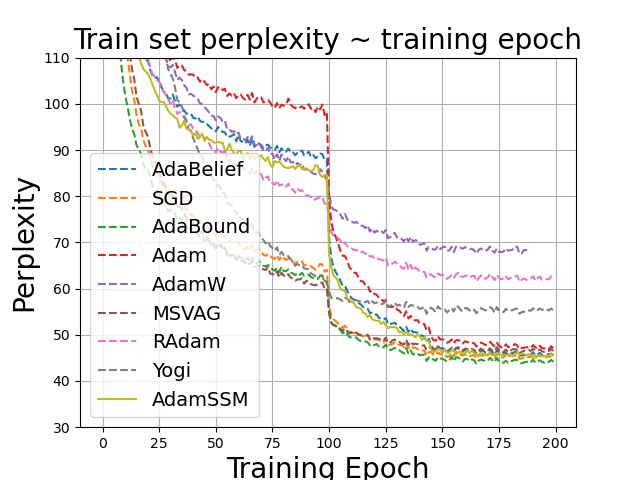}
  \caption{training data}
  \label{fig:train_2layer}
  \end{center}
\end{subfigure}%
\begin{subfigure}{.5\textwidth}
  \begin{center}
  \includegraphics[width = \textwidth]{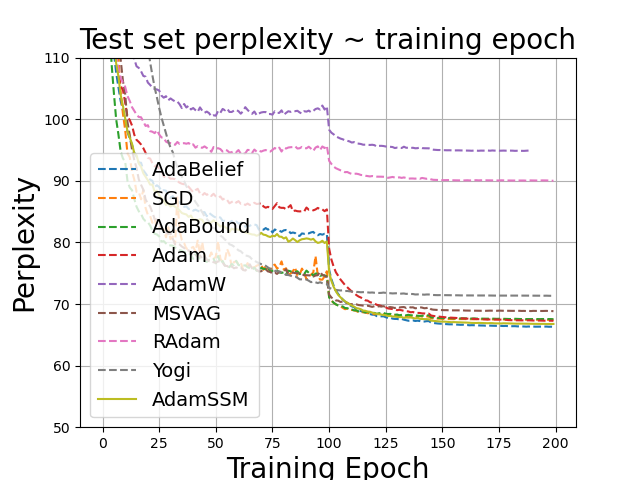}
  \caption{test data}
  \label{fig:test_2layer}
  \end{center}
\end{subfigure}
\caption{\it Accuracy for language modeling task on Penn TreeBank dataset with 2-layer LSTM architecture trained with different algorithms, for (a) training data and (b) test data.}
\label{fig:2layer}
\end{figure*}

\begin{figure*}[htb!]
\centering
\begin{subfigure}{.5\textwidth}
  \begin{center}
  \includegraphics[width = \textwidth]{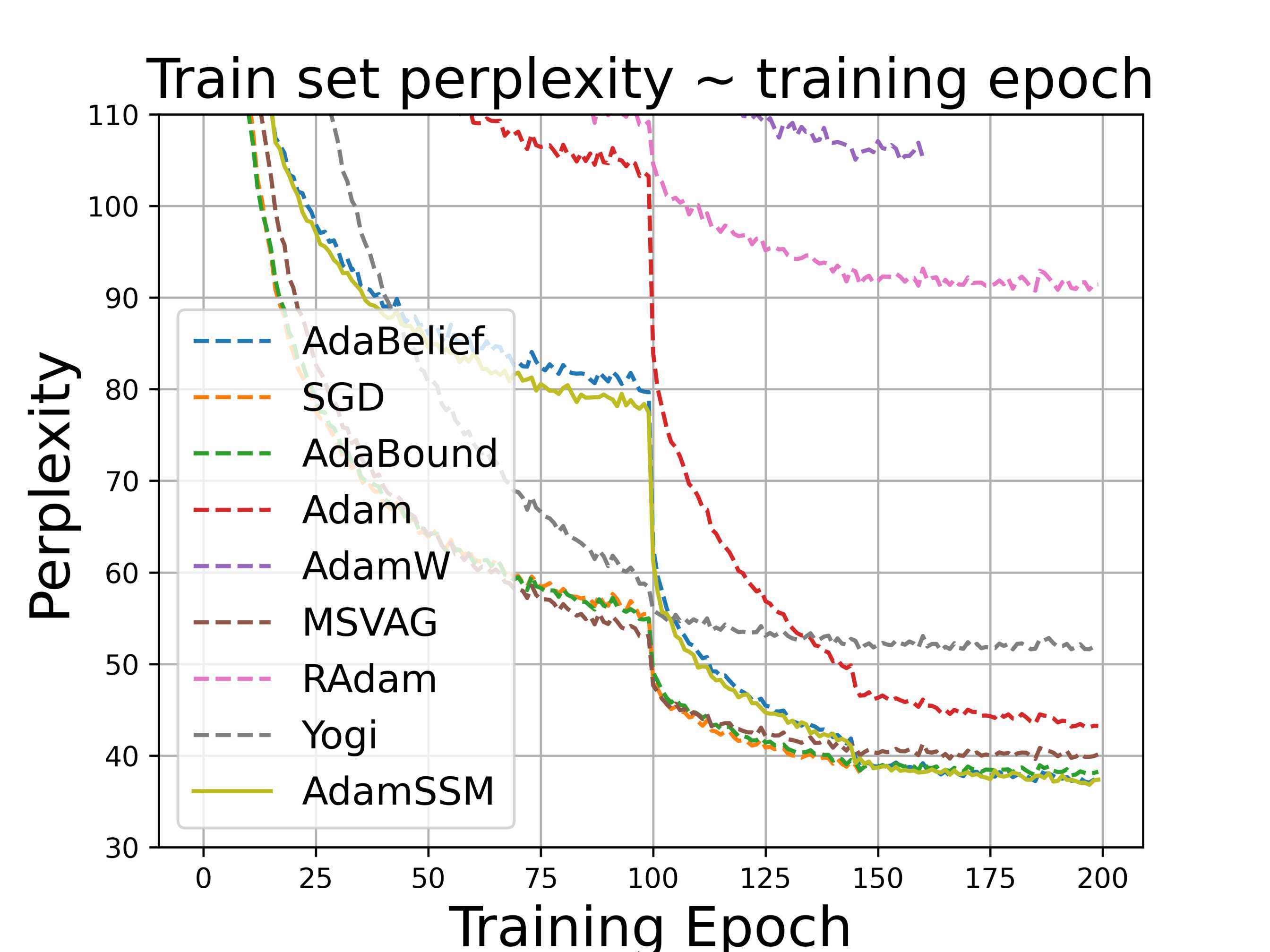}
  \caption{training data}
  \label{fig:train_3layer}
  \end{center}
\end{subfigure}%
\begin{subfigure}{.5\textwidth}
  \begin{center}
  \includegraphics[width = \textwidth]{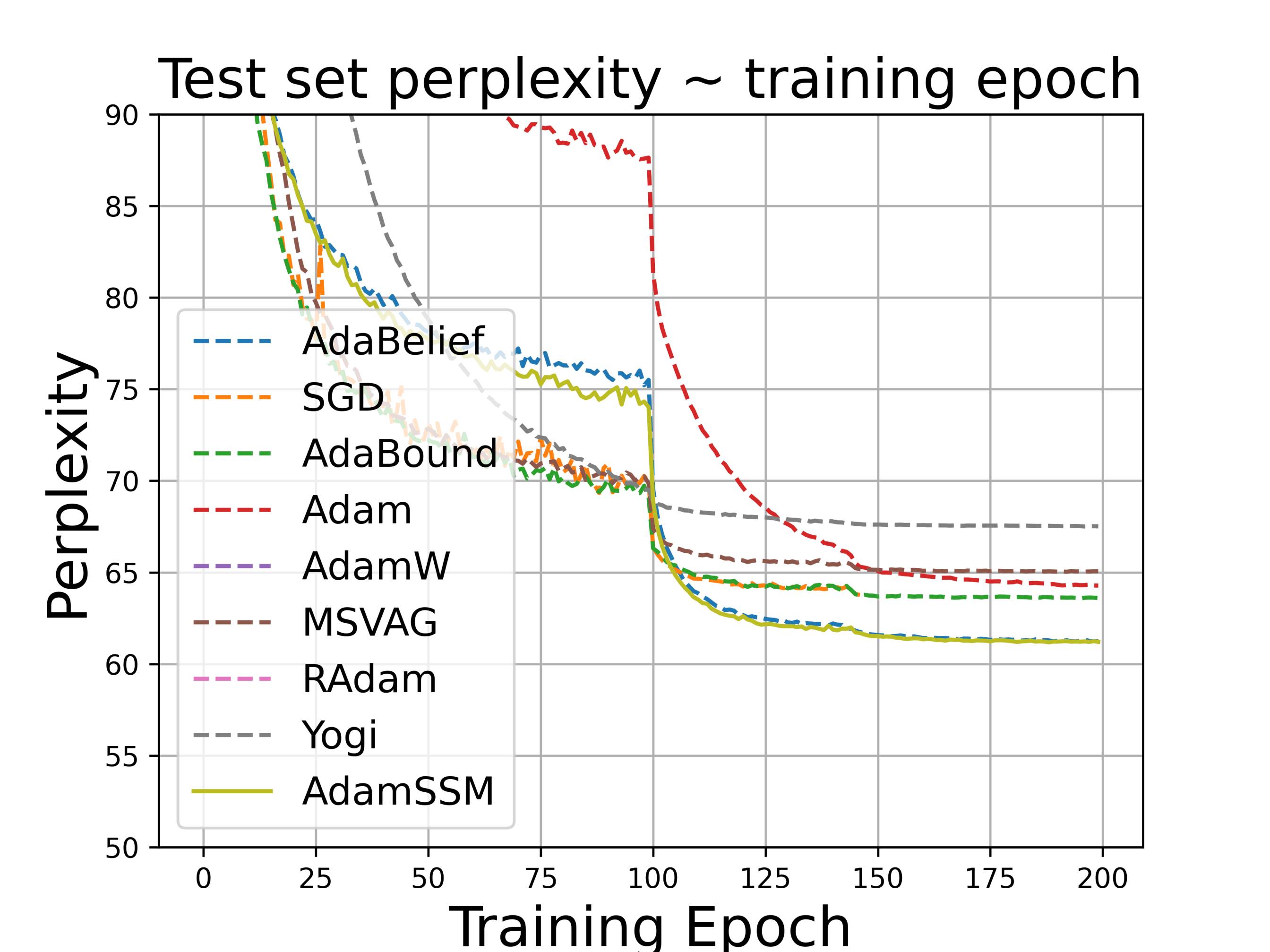}
  \caption{test data}
  \label{fig:test_3layer}
  \end{center}
\end{subfigure}
\caption{\it Accuracy for language modeling task on Penn TreeBank dataset with 3-layer LSTM architecture trained with different algorithms, for (a) training data and (b) test data.}
\label{fig:3layer}
\end{figure*}

\begin{table*}[htb!]
\caption{Comparisons between best training accuracy, best test accuracy, and number of training epochs required to achieve these accuracies for different algorithms on image classification task with ResNet34.}
\begin{center}
\begin{tabular}{|c||c|c||c|c|}
\hline
Training algorithm & Test accuracy & Epoch & Train accuracy & Epoch \\
\hline
\hline
AdaBelief & $95.44$ & $194$ & $99.988$ & $193$ \\
\hline
AdaBound & $94.85$ & $190$ & $99.998$ & $191$ \\
\hline
Adam & $93.02$ & $189$ & $99.308$ & $190$ \\
\hline
AdamW & $94.59$ & $164$ & ${\bf 100.0}$ & $169$ \\
\hline
Fromage & $94.51$ & $165$ & $99.992$ & $165$ \\
\hline
MSVAG & $94.44$ & $199$ & $99.996$ & $185$ \\
\hline
RAdam & $94.33$ & $182$ & ${\bf 100.0}$ & $179$ \\
\hline
SGD & $94.64$ & $155$ & $99.272$ & $169$ \\
\hline
Yogi & $94.71$ & $182$ & $99.972$ & $192$ \\
\hline
AdamSSM (Proposed) & ${\bf 95.61}$ & $174$ & $99.99$ & $188$ \\
\hline
\end{tabular}
\end{center}
\label{tab:resnet_full}
\end{table*}

\begin{table*}[htb!]
\caption{Comparisons between best training accuracy, best test accuracy, and number of training epochs required to achieve these accuracies for different algorithms on image classification task with VGG11.}
\begin{center}
\begin{tabular}{|c||c|c||c|c|}
\hline
Training algorithm & Test accuracy & Epoch & Train accuracy & Epoch \\
\hline
\hline
AdaBelief & $91.41$ & $193$ & $99.784$ & $197$ \\
\hline
AdaBound & $90.62$ & $176$ & $99.914$ & $193$\\
\hline
Adam & $88.40$ & $197$ & $94.028$ & $199$ \\
\hline
AdamW & $89.39$ & $166$ & $99.312$ & $198$ \\
\hline
Fromage & $89.77$ & $162$ & $99.730$ & $170$ \\
\hline
MSVAG & $90.24$ & $187$ & ${\bf 99.948}$ & $192$ \\
\hline
RAdam & $89.30$ & $195$ & $98.984$ & $196$ \\
\hline
SGD & $90.11$ & $188$ & $96.436$ & $195$\\
\hline
Yogi & $90.67$ & $192$ & $99.868$ & $196$ \\
\hline
AdamSSM (Proposed) & ${\bf 91.49}$ & $185$ & $99.792$ & $187$ \\
\hline
\end{tabular}
\end{center}
\label{tab:vgg_full}
\end{table*}

\begin{table*}[htb!]
\caption{Comparisons between best training set perplexity, best test set perplexity, and number of training epochs required to achieve these perplexities for different algorithms on language modeling task with 1-layer LSTM.}
\begin{center}
\begin{tabular}{|c||c|c||c|c|}
\hline
Training algorithm & Test accuracy & Epoch & Train accuracy & Epoch \\
\hline
\hline
AdaBelief & $84.63$ & $199$ & $58.25$ & $192$ \\
\hline
AdaBound & $84.78$ & $199$ & $62.36$ & $155$ \\
\hline
Adam & ${\bf 84.28}$ & $196$ & $58.26$ & $155$ \\
\hline
AdamW & $87.80$ & $194$ & ${\bf 55.33}$ & $155$ \\
\hline
MSVAG & $84.68$ & $199$ & $63.59$ & $167$ \\
\hline
RAdam & $88.57$ & $196$ & $55.81$ & $155$ \\
\hline
SGD & $85.07$ & $199$ & $63.64$ & $155$ \\
\hline
Yogi & $86.59$ & $199$ & $69.22$ & $155$ \\
\hline
AdamSSM (Proposed) & $84.61$ & $199$ & $58.93$ & $192$ \\
\hline
\end{tabular}
\end{center}
\label{tab:lstm1_full}
\end{table*}

\begin{table*}[htb!]
\caption{Comparisons between best training set perplexity, best test set perplexity, and number of training epochs required to achieve these perplexities for different algorithms on language modeling task 2-layer LSTM.}
\begin{center}
\begin{tabular}{|c||c|c||c|c|}
\hline
Training algorithm & Test accuracy & Epoch & Train accuracy & Epoch \\
\hline
\hline
AdaBelief & ${\bf 66.29}$ & $199$ & $45.48$ & $184$ \\
\hline
AdaBound & $67.53$ & $199$ & ${\bf 43.65}$ & $165$ \\
\hline
Adam & $67.27$ & $199$ & $46.86$ & $184$ \\
\hline
AdamW & $94.86$ & $186$ & $67.51$ & $184$ \\
\hline
MSVAG & $68.84$ & $199$ & $45.90$ & $184$ \\
\hline
RAdam & $90.00$ & $199$ & $61.48$ & $184$ \\
\hline
SGD & $67.42$ & $197$ & $44.79$ & $165$ \\
\hline
Yogi & $71.33$ & $199$ & $54.53$ & $143$ \\
\hline
AdamSSM (Proposed) & $66.75$ & $198$ & $44.92$ & $190$ \\
\hline
\end{tabular}
\end{center}
\label{tab:lstm2_full}
\end{table*}

\begin{table*}[htb!]
\caption{Comparisons between best training set perplexity, best test set perplexity, and number of training epochs required to achieve these perplexities for different algorithms on language modeling task 3-layer LSTM.}
\begin{center}
\begin{tabular}{|c||c|c||c|c|}
\hline
Training algorithm & Test accuracy & Epoch & Train accuracy & Epoch \\
\hline
\hline
AdaBelief & $61.24$ & $194$ & $37.06$ & $197$ \\
\hline
AdaBound & $63.58$ & $195$ & $37.85$ & $193$ \\
\hline
Adam & $64.28$ & $199$ & $43.11$ & $197$ \\
\hline
AdamW & $104.49$ & $159$ & $104.94$ & $155$ \\
\hline
MSVAG & $65.04$ & $192$ & $39.64$ & $185$ \\
\hline
RAdam & $93.11$ & $199$ & $90.75$ & $185$ \\
\hline
SGD & $63.77$ & $146$ & $38.11$ & $146$ \\
\hline
Yogi & $67.51$ & $196$ & $51.46$ & $164$ \\
\hline
AdamSSM (Proposed) & ${\bf 61.18}$ & $188$ & ${\bf 36.82}$ & $197$ \\
\hline
\end{tabular}
\end{center}
\label{tab:lstm3_full}
\end{table*}

\end{document}